\documentclass[lettersize,journal]{IEEEtran}
\usepackage{amsmath,amsfonts}
\usepackage{array}
\usepackage{textcomp}
\usepackage{stfloats}
\usepackage{float}
\usepackage{subcaption}
\usepackage{url}
\usepackage{verbatim}
\usepackage{graphicx}
\usepackage{tabularx}
\usepackage{multirow}
\usepackage{amssymb}
\usepackage[dvipsnames]{xcolor}
\usepackage{cite}
\usepackage[ruled,vlined,linesnumbered]{algorithm2e}
\usepackage{booktabs}
\usepackage{multirow}
\usepackage{todonotes}
\usepackage[
    pagebackref=true, 
	colorlinks=true,
	linkcolor=blue, 
	citecolor=red, 
	urlcolor=blue, 
]{hyperref}

\usepackage{orcidlink}

\begin{document}

\title{Human-Centric Perception for Child\\ Sexual Abuse Imagery}

\author{Camila Laranjeira\orcidlink{0000-0002-0521-0432}, João Macedo\orcidlink{0000-0001-5558-3088}, Sandra Avila\orcidlink{0000-0001-9068-938X}, Fabrício Benevenuto\orcidlink{https://orcid.org/0000-0001-6875-6259}, Jefersson A. dos Santos\orcidlink{0000-0002-8889-1586},~\IEEEmembership{Senior Member,~IEEE}
\thanks{Disclaimer: This paper displays blurred images of adult pornography. Reader discretion is advised.}
\thanks{C. Laranjeira was with Universidade Federal de Minas Gerais (UFMG), Brazil.}
\thanks{J. Macedo is with UFMG, Brazil, and the Federal Police of Brazil.}
\thanks{F. Benevenuto is with UFMG, Brazil.}
\thanks{S. Avila is with Universidade Estadual de Campinas (UNICAMP), Brazil.}
\thanks{J. A. dos Santos is with the University of Sheffield, United Kingdom.}
}

\markboth{IEEE TRANSACTIONS ON INFORMATION FORENSICS AND SECURITY}%
{Laranjeira \MakeLowercase{\textit{et al.}}: Human-Centric Perception for Child Sexual Abuse Imagery}


\maketitle

\begin{abstract}
Law enforcement agencies and non-gonvernmental organizations handling reports of Child Sexual Abuse Imagery (CSAI) are overwhelmed by large volumes of data, requiring the aid of automation tools. However, defining sexual abuse in images of children is inherently challenging, encompassing sexually explicit activities and hints of sexuality conveyed by the individual's pose, or their attire. 
CSAI classification methods often rely on black-box approaches, targeting broad and abstract concepts such as pornography.
Thus, our work is an in-depth exploration of tasks from the literature on Human-Centric Perception, across the domains of safe images, adult pornography, and CSAI, focusing on targets that enable more objective and explainable pipelines for CSAI classification in the future.
We introduce the Body-Keypoint-Part Dataset (BKPD), gathering images of people from varying age groups and sexual explicitness to approximate the domain of CSAI, along with manually curated hierarchically structured labels for skeletal keypoints and bounding boxes for person and body parts, including head, chest, hip, and hands. We propose two methods, namely BKP-Association and \text{YOLO-BKP}, for simultaneous pose estimation and detection, with targets associated per individual for a comprehensive decomposed representation of each person. 
Our methods are benchmarked on COCO-Keypoints and COCO-HumanParts, as well as our human-centric dataset, achieving competitive results with models that jointly perform all tasks. Cross-domain ablation studies on BKPD and a case study on RCPD highlight the challenges posed by sexually explicit domains. Our study addresses previously unexplored targets in the CSAI domain, paving the way for novel research opportunities. 

\end{abstract}


\section{Introduction}

Child Sexual Abuse Material (CSAM) is defined under Brazilian law as ``sexually explicit or pornographic scene involving a child or adolescent''~\cite{lei_11829_2008}. The Sexual Offences Guideline~\cite{ministry_of_justice_2013} from the United Kingdom offers a detailed classification, categorizing materials from less severe, such as erotic posing, to more severe, including sadism. The terminology may vary, as reflected by these legislations, referring to it respectively as Child Pornography and Indecent Images of Children. We adopt the term CSAM following the Luxembourg Guidelines~\cite{ecpat_luxembourg_2016}. We may also use Child Sexual Abuse Imagery (CSAI) to emphasize our focus on images.



Research on CSAI often leverages seized hard drives or investigation reports, assigning positive CSAI labels to all images~\cite{macedo2025child}. However, law enforcement agents (LEA) rely on the investigation context, including additional data or images. Thus, such datasets may contain a wide variety of what can be considered indecent, lacking an objective definition. To circumvent this issue, Macedo et al.~\cite{macedo2018benchmark} introduced the Region-based Child Pornography Dataset (RCPD), a manually curated benchmark annotating age groups and bounding boxes for nude private parts. It enables an objective, though limited, definition of CSAI as images with nudity and children. 

Concepts such as pornography and indecency often lead to disagreements among annotators~\cite{dalins2018laying} and difficulties in task definition~\cite{avila2013pooling}. These tasks have faced criticism from scholars in pornography studies~\cite{gehl2017training}, arguing that the computer vision community's understanding of pornography is biased, whether in hand-crafting features or curating datasets. 
To think about CSA more objectively, we refer to the categorization of apprehended images by child protection institutions. A report by the Internet Watch Foundation~\cite{iwf2022report} discloses metadata assigned by forensic experts, with concepts such as nudity, posing, or sexually explicit activities. Although nudity has been used as a feature for CSAI, neither pose nor activity were investigated in practice, despite being mentioned by LEA~\cite{kloess2019challenges}. 

Towards this goal, we explore tasks within the broad umbrella of human-centric perception (HCP)~\cite{tang2023humanbench}, specifically: pose estimation, person detection, and detection of body parts commonly involved in sexual activities. 
A decomposed representation of human bodies enables higher-level tasks, such as recognizing human actions and interactions~\cite{feng2022skeleton}. 
Our work explores in depth the behavior of models targeting the aforementioned HCP tasks across the domains of safe images, adult pornography, and CSAI, surfacing existing challenges in sexually explicit contexts that must be addressed before relying on these targets for downstream tasks and, ultimately, CSAI classification. Our long-term goal is to shift the focus of the CSAI literature away from abstract definitions of pornography, focusing on objective and more explainable targets.



In practice, we build on YOLO-Pose~\cite{maji2022yolo}, a model trained on COCO~\cite{lin2014microsoft} for person detection and pose estimation inherently associated per individual. We propose two approaches to incorporate the detection of human parts. The first, BKP-Association, inspired by Zhou et al.~\cite{zhou2024bpjdet}, detects bodies and parts independently, later associating them with our proposed skeleton-based algorithm. 
The second, YOLO-BKP, adds a novel detection head for body parts into YOLO-Pose, enabling end-to-end inference of associated targets. 

We introduce the Body-Keypoint-Part Dataset (BKPD), with samples from Open Images~\cite{kuznetsova2018open} representing safe images of people and sexually explicit images from the Sexual Organ Detection (SOD) dataset~\cite{tabone2021pornographic}. The novelty lies in the released labels, with skeleton keypoints and hierarchical bounding boxes for people and body parts. To our knowledge, there is no such dataset leveraging samples from adult pornography. 
 
We apply our pipeline to real CSAI, leveraging RCPD to extract actionable insights and highlight specific domain challenges.
In summary, this work has the following contributions:
\begin{itemize}
    \item The Body-Keypoint-Part Dataset (BKPD), with manually labeled legal images that approximate the CSAI domain;
    \item Two methods: a post-processing algorithm for target association (BKP-Association) and an end-to-end pipeline to infer inherently associated targets (YOLO-BKP);
    \item Actionable insights derived from real CSAI.
\end{itemize}

Section~\ref{sec:rwork} provides a literature review of CSAI classification and human-centric perception. Section~\ref{sec:dataset} outlines the characteristics of our BKP-Dataset. Section~\ref{sec:method} describes the proposed methodologies. Lastly, Section~\ref{sec:experiments} benchmarks our proposition and presents key findings from the CSAI domain.

\section{Related Work}
\label{sec:rwork}

\subsection{CSAI Classification}
\label{sec:rwork-csai}



Table~\ref{tab:mc-methods} summarizes the history of CSAI classification. Most works focus mainly on two tasks: age estimation~\cite{sae2014towards, dalins2018laying, macedo2018benchmark, gangwar2021attm, rondeau2022deep}
and concepts related to indecency such as skin exposure~\cite{de2010nudetective, peersman2016icop}, nudity~\cite{tabone2021pornographic} or pornography~\cite{vitorino2018leveraging, macedo2018benchmark, gangwar2021attm}. The table is divided between main tasks, i.e., features or models proposed by the authors, and auxiliary tasks, such as off-the-shelf pre-processing. Regarding auxiliary tasks, while indecency leveraged skin segmentation in the past, researchers nowadays propose end-to-end training. Face detection is the only auxiliary task utilized to support age estimation.  

\begin{table}[t]
    \centering
    \begin{tabularx}{\linewidth}{clll}
        \toprule
        \textbf{Reference} & \textbf{Year}   & \textbf{Main Task} & \textbf{Auxiliary Task} \\ 
        \midrule
        \multirow{1}{*}{\cite{de2010nudetective}} & 2010 & Nudity Classification & Skin Segmentation \\ \midrule
        \multirow{1}{*}{\cite{ulges2011automatic}} & 2011 & CSAI Classification &   \\ \midrule
        \multirow{1}{*}{\cite{schulze2014automatic}} & 2014 & CSAI Classification & Skin Segmentation \\ \midrule
        \multirow{2}{*}{\cite{sae2014towards}} & \multirow{2}{*}{2014}
         & Nudity Classification & Skin Segmentation  \\
         & & Child Classification & Face Detection  \\ \midrule
         \multirow{1}{*}{\cite{peersman2016icop}} & 2016 & CSAI Classification & Skin Segmentation \\ \midrule
         \multirow{1}{*}{\cite{yiallourou2017detection}} & 2017 & Appropriateness Classification & Refer to Table \ref{tab:dc-methods} \\ \midrule
         \multirow{2}{*}{\cite{vitorino2018leveraging}} & \multirow{2}{*}{2018} & CSAI Classification &  \\
         & & Pornography Classification & \\ \midrule
         \multirow{4}{*}{\cite{macedo2018benchmark}} & \multirow{4}{*}{2018} & Age Classification & Face Detection \\
         & & Child Classification & Face Detection \\
         & & Gender Classification & Face Detection \\
         & & Pornography Classification &  \\ \midrule
         \multirow{3}{*}{\cite{dalins2018laying}} & \multirow{ 3}{*}{2018} & Child Classification &  \\
         & & Pornography Classification &  \\
         & & CSAI Severity Classification &  \\ \midrule
         \multirow{2}{*}{\cite{gangwar2021attm}} & \multirow{ 2}{*}{2021} & Age Classification & Face Detection \\
         & & Pornography Classification & \\ \midrule
         \multirow{2}{*}{\cite{tabone2021pornographic}} & \multirow{2}{*}{2021} & Sexual Organ Detection &  \\
         & & Pornography Classification & \\ \midrule
         \multirow{2}{*}{\cite{rondeau2022deep}} & \multirow{2}{*}{2022} & Age Classification & Face Detection  \\
         & & Pornography Classification & \\ 
        \bottomrule
    \end{tabularx}
    \caption{Overview of CSAI classification methods. Main tasks are features or models proposed by the authors, while auxiliary tasks involve off-the-shelf pre-processing methods.}
    \label{tab:mc-methods}
\end{table}


Regarding age estimation for CSAI, research often involves collecting child images. Anda et al.~\cite{anda2020deepuage} introduce VisAGe, a dataset comprising \(\sim \)19k faces of individuals under 18 from creative-commons Flickr images. Similarly, \cite{gangwar2021attm} created Juvenile-80k, compiling  \(\sim \)24k underage images from age estimation datasets and search engines. This collection, also employed by many other works~\cite{chaves2020improving, anda2021vec2uage, roopak2023comparison}, aims to address the child/adult data imbalance, highlighting the lack of effective methods for underage age estimation. However, it disregards ethical standards for children's biometric data collection.

To model indecency, early methods proposed low and middle-level features for skin exposure as a proxy for nudity~\cite{de2010nudetective, peersman2014icop}, while recent approaches follow two main paths: (1) training end-to-end models for adult pornography classification~\cite{macedo2018benchmark, vitorino2018leveraging, gangwar2021attm}, or (2) framing the task as detection of nude private parts~\cite{tabone2021pornographic}. Regarding the first approach, Yahoo's OpenNSFW model~\cite{mahadeokar2016open} has been widely used off-the-shelf for CSAI classification~\cite{gangwar2017pornography, macedo2018benchmark, rondeau2022deep}. Notably, \text{Macedo et al.~\cite{macedo2018benchmark}} suggests lowering the pornography classification threshold down to $\tau=0.1$ to increase recall on child-related images, indicating a disparity between domains. Few works \cite{vitorino2018leveraging,gangwar2021attm} train their own classifiers, relying on datasets to accurately represent the pornography domain --- a problematic assumption given the unclear target definition~\cite{avila2013pooling, dalins2018laying} and minimal attention to data collection~\cite{gangwar2021attm}. Automated large-scale data collection, as seen in Gangwar et al.~\cite{gangwar2021attm}, raises ethical concerns 
and often results in low-quality datasets due to insufficient curation, offering no guarantees regarding the representativeness of the targeted domain.

Conversely, focusing on nude body parts revisits the limitations of earlier methods, equating nudity with pornography. 
Macedo et al.~\cite{macedo2018benchmark} propose RCPD, a dataset with manually label private parts --- chest, buttocks, and genitals --- leveraging the nudity as a definition for indecency. Tabone et al.~\cite{tabone2021pornographic} propose to detect similar targets, along with interaction targets like coitus and objects like sex toys. However, they reported a significant performance drop when applying models trained on adult images to CSAI; for instance, the ``male genital'' class declined from $91\%$ recall for adults to below $65\%$ for children. Forensic experts also state that CSAI does not require nudity, with clothing used as props to infantilize or increase the perception of sexuality~\cite{kloess2019challenges}.

From the works outlined in Table~\ref{tab:mc-methods}, Yiallourou et al.~\cite{yiallourou2017detection} stand out with a data-centric view. The authors propose a synthetic dataset associating levels of appropriateness according to several features, such as gestures, scenes, and facial expressions. Table~\ref{tab:dc-methods} highlights other data-centric works, proposing labeling schemes for CSAI datasets. While model-centric approaches are limited in pursued tasks, forensic~experts tend towards a broader perspective on targets. For instance, Majura~\cite{dalins2018laying} is a comprehensive labeling schema to classify CSAI severity according to the Child Exploitation Tracking System (CETS), a guideline for file annotation in investigations.

\begin{table}[t]
    \centering
    \setlength\tabcolsep{0.5pt}
    \begin{tabular*}{\linewidth}{@{\extracolsep{\fill}} cll}
        \toprule
        \textbf{Reference} & \textbf{Year} & \textbf{Targets (Task)} \\ \midrule
        \cite{yiallourou2017detection} & 2017 & 
        \begin{tabular}[x]{@{}l@{}}
            Person (C), Pose (C), Action (C), Age (C), Scene (C),\\
            Attire (C), Face expression (C), Illumination (C).
        \end{tabular} \\
        \midrule
        RCPD~\cite{macedo2018benchmark} & 2018 &
        \begin{tabular}[x]{@{}l@{}}
            Person (D), Face (D), Nude private parts (D),\\
            Age (C), Gender (C).
        \end{tabular} \\
        \midrule
        Majura~\cite{dalins2018laying} & 2018 &
        \begin{tabular}[x]{@{}l@{}}
            Pornography (C), Nudity (C), Participants (C),\\
            Penetration (C), Props (C), Virtual (C), BDSM (C),\\
            Bodily Fluids (C).
        \end{tabular} \\ \bottomrule
    \end{tabular*}
    \caption{Overview of CSAI labeling schemes in the literature. C: Classification, D: Detection.}
    \label{tab:dc-methods}
\end{table}

        
        

A deeper exploration of law enforcement's perspective is the work of Kloess et al.~\cite{kloess2019challenges}. Although they do not attempt automatic classification, the authors draw insights from CSAI experts. They highlight visual cues that may cause or solve ambiguities in classification, such as a person's pose, clothing, or the background. They decompose the target problem into diverse tasks, making classification less subjective. 
Similarly, our work does not attempt direct CSAI classification; instead, it explores the challenges of related human-centric subtasks.

\subsection{Human-Centric Perception}
\label{sec:rwork-hcp}

Human-Centric Perception (HCP) is a collection of computer vision tasks focused on humans, uniting a wide range of downstream applications. Our purpose is to investigate human-centered auxiliary tasks related to the CSAI domain. A benchmark named \textit{HumanBench}~\cite{tang2023humanbench} gathers six tasks from dozens of HCP datasets into a comprehensive evaluation set, namely person detection, pose estimation, human parsing, attribute recognition, crowd counting, and person re-identification.

Through the lenses of CSAI classification, we disregard crowd counting and person re-identification as they are unrelated to our goal. Crowd counting focuses on large groups, while CSA images typically involve fewer individuals~\cite{laranjeira2022seeing}. Person re-identification, i.e., matching identities across references, could locate victims in a perpetrator's database but is irrelevant for identifying visual cues of abuse. Furthermore, attribute recognition, common in pedestrian monitoring, targets urban-specific targets like ``talking on the phone"~\cite{liu2017hydraplus}. Modeling it for CSAI classification requires a deeper semantic understanding of the domain, which is currently limited~\cite{laranjeira2022seeing}.

The tasks of person detection, pose estimation, and human parsing are closely tied to our goal. Person detection is a fundamental premise in CSAI, as images without people can be disregarded. A survey~\cite{zaidi2022survey} compares modern object detection methods, with YOLOv4 and YOLOv5 excelling in precision and processing speed. This is ideal for law enforcement, which handles large datasets with limited resources~\cite{sanchez2019practitioner}. 

Regarding pose, it has been mentioned as a relevant attribute for CSAI~\cite{yiallourou2017detection, kloess2019challenges}, used by LEA to disambiguate images. Pose is also a valuable input for other tasks such as age estimation~\cite{korban2023taa} and action recognition~\cite{feng2022skeleton}. To our knowledge, pose estimation has never been explored in practice for the CSAI domain. YOLO-Pose~\cite{maji2022yolo}, built on YOLOv5 and trained for joint pose estimation person detection, achieves remarkable results on person-centric images. YOLO-Pose was later updated to YOLOv7~\cite{wang2023yolov7}, further improving performance. It serves as a starting point for our proposition.

Lastly, human parsing is traditionally defined by dense pixel-wise labeling of body parts or garments. However, it has been approached as a detection task, decomposing the body into semantically meaningful parts using cheaper labels and lighter models. Two key approaches stand out~\cite{yang2023deep}: hierarchical and multi-task. The former relates to the inherent hierarchy of body parts, which has been successfully translated into hierarchical models. 
Conversely, multi-task human parsing is associated to pose estimation, as there is an intrinsic relationship between skeleton coordinates and body parts~\cite{yang2023deep}, which can boost both tasks in a simultaneous learning~setting.

Human parsing offers a novel approach to a key substitute task for CSAI classification: locating private body parts. Prior works~\cite{tabone2021pornographic, macedo2018benchmark} focus on distinguishing between clothed and nude parts and even demographic attributes like gender expression. In contrast, human parsing assumes universal commonalities across human bodies, enabling accurate parsing regardless of age, gender, or nudity. 

A recent work, entitled BPJDet~\cite{zhou2024bpjdet}, enhances YOLO to jointly detect bodies and human parts, introducing an association decoding algorithm to assign parts to the corresponding person. The authors extend YOLO's traditional detection target for the person class, encompassing $(x, y)$ coordinates for all body parts in a single vector. These coordinates are then used by their association algorithm. BPJDet achieves superior results with a more efficient, near real-time approach.

HCP tasks are under-explored for the CSAI domain, thus there is little knowledge on domain-specific challenges that might hinder future CSAI classification approaches.

\section{BKP-Dataset}
\label{sec:dataset}
We propose a novel dataset comprised of human-centric images, and joint labels for each individual, including skeleton keypoints and bounding boxes for the person's body and the following human parts: head, chest, hip, left hand, and right hand. We aim at targets present in all human bodies, regardless of sexual explicitness or demographic dimensions such as age and gender expression, thus robust to domain shifts.


The need for a novel dataset has several reasons. Firstly, we approximate the domain of CSAI, including safe images of people from 
varying
age groups, along with sexually explicit images of adults. We propose novel detection targets---chest and hip---not addressed by other datasets. Lastly, our goal to infer associated targets reinforces the need for a database providing joint labels. Henceforth, we refer to our proposed dataset as \textit{BKPD}, referencing their targets: \textit{Body}, \textit{Keypoints}, and \textit{Parts}, with the acronym ending with \textit{D} for \textit{Dataset}. This section details the following characteristics of BKPD:

\begin{itemize}
    \item \textbf{Novel detection targets}. Body parts forensic experts consider relevant. 
    \item \textbf{Diversity of characteristics} relevant to the CSAI domain, namely age and sexual explicitness.
    \item \textbf{Hierarchic association} of labels, with body, keypoints, and part boxes associated to a single entity.
    \item \textbf{Manual labeling}, ensuring the quality of targets.
\end{itemize}

Regarding pose keypoints, we follow the COCO-Keypoints representation, since it leverages one of the largest and widely used benchmarks. It includes five keypoints for the head, six for the arms (shoulders, elbows, and wrists), and six for the legs (hips, knees, and ankles). Furthermore, the choice of body parts follows from our literature review. Locating the face is often crucial for demographic targets, such as age. However, we focus on heads, which may not always include visible faces. If needed, face visibility can be inferred from facial keypoints. Head localization, regardless of face, is useful for detecting purposeful occlusions or modeling sexual activities as relations of body parts.

Regarding chest and hip, we draw inspiration from the literature. Macedo et al.~\cite{macedo2018benchmark} provide detection labels only for nude private parts in a CSAI dataset, while \text{Tabone et al.~\cite{tabone2021pornographic}} propose a YOLO-based approach to automate similar targets, likewise limited to nudity. As Gehl et al.~\cite{gehl2017training} argue, it is shortsighted to equate sexuality with nudity. For instance, a photograph focused on the hip of a child should be retrieved as a potential CSAI regardless of full nudity. Lastly, we propose adding hands as a target, given the documented activities in reported CSAI~\cite{iwf2022report}, in which most images are labeled as non-penetrative activities, including touching or rubbing.


To approximate the domain of CSAI in terms of age and sexuality, we gathered samples from two datasets: \textbf{Open Images}\cite{kuznetsova2018open} for safe images of people and \textbf{SOD} (Sexual Organ Detection)\cite{tabone2021pornographic} for explicit content. SOD includes detection labels for sensitive regions and actions, divided into: coitus, anal, female genitalia open, female genitalia posing, female breast, female buttock, male genitalia, male buttock, and sex toy. Open Images, a large-scale dataset, provides detection labels for people categorized by perceived age and binary gender through the labels woman, man, girl, and boy. 

From the Open Images dataset, we selected 300 images from each selected category, ensuring diversity across age groups. Images with more than seven individuals were excluded, as this was the maximum quantity observed in a real CSAI dataset~\cite{laranjeira2022seeing}. It resulted in 983 unique samples. For the SOD dataset, we randomly selected approximately 1,000 samples. The final dataset comprises a total of 1,985 samples.

\begin{figure}[t]
    \centering
    \includegraphics[width=\linewidth]{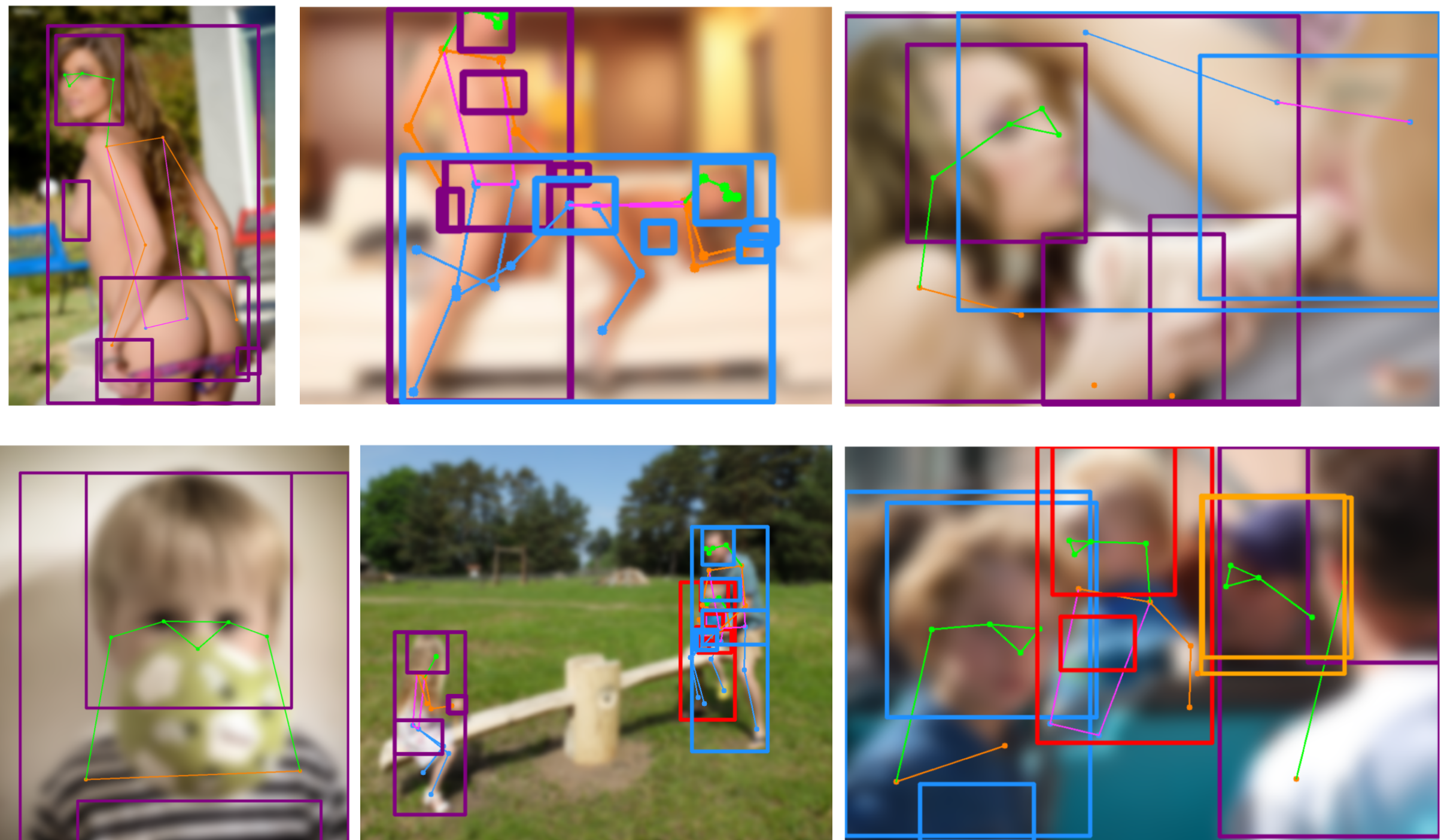}
    \caption[Examples of labeled samples from BKPD.]{Examples of labeled samples from BKPD. Top row: sourced from SOD, bottom row: sourced from Open Images. Boxes are color-coded per associated individual.}
    \label{fig:bkpd-samples}
\end{figure}

We adopted a semiautomatic labeling approach, starting from readily available targets to expedite the laborious manual labeling process. Open Images provides non-exhaustive detection targets for persons, heads, and hands, directly equivalent to BKPD classes. SOD targets include buttocks and genitalia (hip regions) as well as female breasts (chest). To supplement missing targets, YOLO-Pose trained on COCO-Keypoints provided person detection and skeleton coordinates. From the latter, we automatically generated low-quality targets for body parts, e.g., head boxes around facial keypoints, chest around shoulders, and hands around wrists. The aforementioned sum of targets was loaded into Label Studio~\cite{labelstudio} to initiate the authors' manual annotation process, which included not only fixing low-quality labels, but also organizing them into a hierarchical structure, thereby associating labels from the same individual. Figure~\ref{fig:bkpd-samples} depicts a few examples from BKPD.

Each sample was labeled and revised by at least two authors, who previously studied the annotation behavior of the COCO-based datasets leveraged by our work. Skeletal keypoints were considered as the center of each keypoint's corresponding joint, while detection targets tightly encompassed the visible corresponding region. While it was straightforward to define the boundaries for head and hands, the authors defined chest as the region from the armpit to the base of the chest, and hip from the hipbones to the inferior gluteus base.

Table \ref{tab:num-annot} summarizes the total bounding boxes annotated. 
Open Images has significantly more person annotations, reflecting a higher number of people per image, with heads being particularly prevalent. In contrast, hips are the most frequent category in SOD, aligning with known CSAI statistics~\cite{laranjeira2022seeing}, where heads and genitals are predominant.

\begin{table}[t]
    \centering
    \begin{tabular}{l|rrrrrr}
        \toprule
        \textbf{Dataset} & \textbf{person} & \textbf{head} & \textbf{chest} & \textbf{hip} & \textbf{l-hand} & \textbf{r-hand} \\ \toprule
        SOD         & 1121 & 951 & 833 & 1013 & 767 & 815 \\ 
        Open-Images & 2303 & 2180 & 1693 & 1238 & 1199 & 1274 \\ \midrule
        Total & 3424 & 3131 & 2526 & 2251 & 1966 & 2089 \\ \bottomrule
    \end{tabular}
    \caption{Total number of annotated parts on BKPD.}
    \label{tab:num-annot}
\end{table}

\section{Methodologies}
\label{sec:method}

We infer associated targets including skeleton keypoints and bounding boxes for person and parts. Building on YOLO-Pose~\cite{maji2022yolo}, we propose two methods: the first, BKP association (Section~\ref{sec:bkp-assoc}), detects bodies and parts independently, later using skeleton information for association. The second, YOLO-BKP (Section~\ref{sec:yolo-bkp}), adds a novel detection head for body parts, achieving end-to-end inference and association.

\subsection{BKP Association}
\label{sec:bkp-assoc}

YOLO-Pose is able to inherently associate keypoints to a person detection by adding a pose estimation head with the same dimensionality of the detection head. Keypoint targets are aligned to the same level, anchor and grid cell of the detection corresponding to the same individual, thus providing the association of both targets at inference time. 


Our first proposal assumes that the most straightforward way of introducing body part detection into YOLO-Pose is to perform multiclass object detection within the existing pipeline, predicting a person's body and its parts independently at the detection head. This, however, requires a further association step in which all predicted boxes for the class person are matched with the corresponding part boxes.

We propose an association algorithm inspired by the simple and effective proposition of Zhou et al.~\cite{zhou2024bpjdet}, while reducing redundancy. For clarity, the authors extended YOLO's detection capabilities by predicting the $(x, y)$ center of each mapped body part alongside the bounding box of the person. In other words, 
given $n$ classes of $n-1$ body parts and the class person, 
the targets for person are accompanied by $x_i, y_i$ coordinates for each associated part, a vector in the form $(x, y, w, h, conf, c_0=1, c_1=0, ..., c_n=0, x_{0}, y_{0}, ... x_{p}, y_{p},)$, 
with $conf$ being the person bounding box prediction confidence and $c_i$ the classification probability for each class $i$. 
Targets for body parts, on the other hand, have no extra prediction effort, producing an output in the form $(x, y, w, h, conf, c_0=0, c_1=1, ..., c_n=0, 0, 0, ... 0, 0,)$. This method's association between person and body parts relies on comparing redundant predictions of $(x, y)$ coordinates for body parts, the one predicted along with the person, and the actual part detection.

Our proposal leverages a pipeline that jointly performs detection and pose estimation. The key insight is that body parts can be effectively associated with the corresponding person by comparing their predicted bounding box center $(x, y)$ with the nearest keypoint cluster. For instance, head detections can be associated with the centroid of face keypoints, while hand detections can be matched to wrist keypoints, following an intuitive body-keypoint-part relationship. The reference work~\cite{zhou2024bpjdet} offers no utility for the redundant prediction of parts. In contrast, pose estimation is inherently valuable for our final application of human-centric perception for CSAI.


All candidate detections output by YOLO go through a non-maximum suppression (NMS) step to remove redundant detections. This process requires defining two hyperparameters, $\tau_{conf}$ to filter out low confidence boxes, and $\tau_{iou}$ to determine whether a pair of boxes should be assessed as redundant. As in Zhou et al.~\cite{zhou2024bpjdet}, we leverage different body and parts thresholds to account for each task's complexity. 


In possession of both the detections and keypoint estimations, it is time to associate part detections to their respective body. Algorithm~\ref{alg:association} presents a simplified version of our proposed BKP association. Let us first define the main variables:

\begin{itemize}
    \item $C^p$: Part classes of the dataset (e.g., head, chest).
    \item $\mathbf{B}$: set of $nb$ detected bodies, already associated to the corresponding keypoints. Each $b \in B$ is in the form ($x, y, w, h, conf, x_0, y_0, v_0, ..., x_k, y_k, v_k$).
    \item $\mathbf{K}$: Centroids $(x_k, y_k)$ of keypoints associated with a part class, calculated from the set $B$ of bodies detected. $K$ is $nb \times nc^p \times 2$, respectively the number of detected bodies $B$, number of part classes $C^p$, and 2D coordinates.
    \item $\mathbf{P}$: set of $np$ detected parts, each $p \in P$ in the form ($x, y, w, h, conf, class$).
\end{itemize}

\begin{algorithm}[t]
\caption{BKP-Association}\label{alg:association}

\For{$c \in C^p$}{
    $K_{c} \gets \text{centroid of keypoints}(B, c)$\;
}
\For{$p \in P$}{
    $(x_p, y_p) \gets (x, y) \text{ of } p$\;
    $c \gets class \text{ of } p$\;
    $closest \gets \arg\min_{k \in K_{c}} \| (x_p, y_p) - (x_k, y_k) \|$\;
    $B_{closest} \gets B_{closest} \cup p$\;
}
\Return $B$\
\end{algorithm}


The process begins by calculating keypoint centroids $K_c$ for each part class across all detected bodies. It proceeds to iterate over the set of detected parts $P$ to find the $closest$ keypoint centroid to the part's coordinates. 
The result is an updated set of detected bodies, joining the corresponding parts.

\subsection{YOLO-BKP}
\label{sec:yolo-bkp}

\begin{figure}[t]
    \centering
    \includegraphics[width=\linewidth]{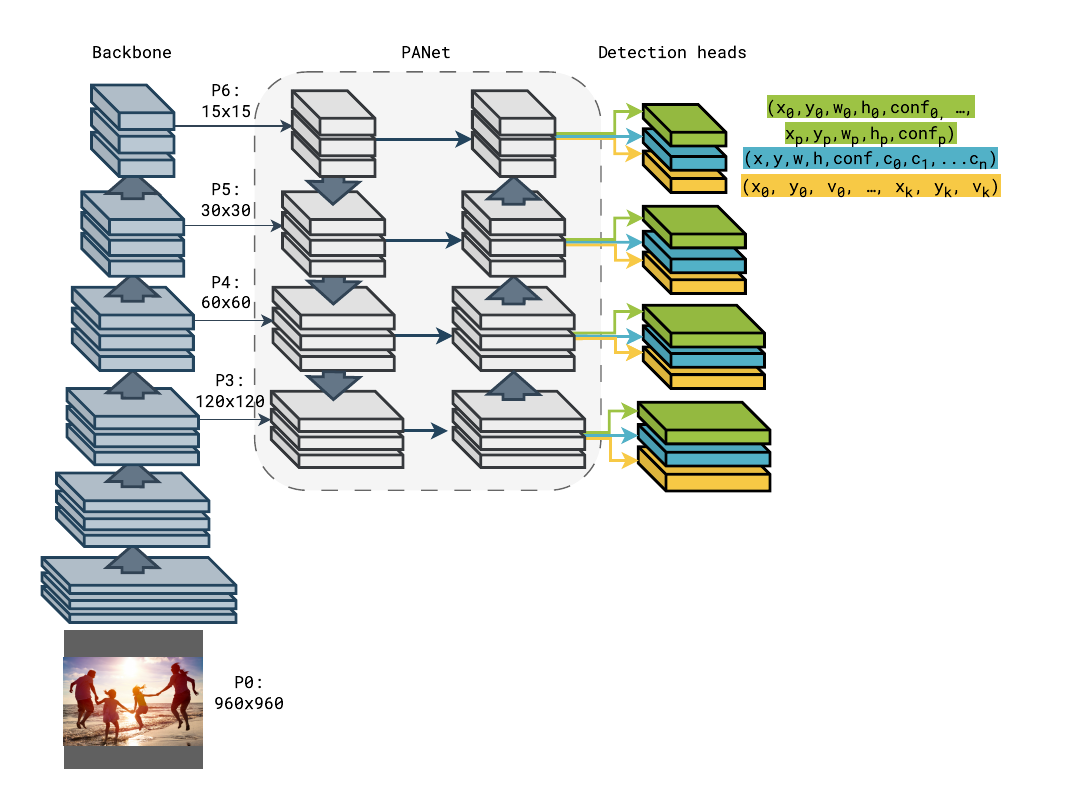}
    \caption{YOLO-BKP architecture.}
    \label{fig:yolo-bkp-arch}
\end{figure}

This proposal builds on YOLO-Pose's joint inference approach by introducing an additional detection head for human parts, as shown in Figure~\ref{fig:yolo-bkp-arch}. Similar to YOLO-Pose's keypoint head, we infer a vector of size $5p$ with $p$ being the number of mapped parts. Each part $i$ is predicted by $5$ components $(x_i,y_i,w_i,h_i,conf_i)$, with categories inferred from their position in the vector. Aligning body, keypoint, and part targets to the same level, anchor, and grid cell of the architecture enables end-to-end prediction of inherently associated targets.

In YOLO’s framework, when detection targets are associated to a grid cell, their $x, y$ coordinates are relative to the cell's position, lying within a $[-0.5, 1.5]$ interval. In other words, the bounding box center is around that cell. For the keypoints in YOLO-Pose, $x, y$ values do not lie within the same interval, i.e., keypoints can be further away from the cell. For YOLO-BKP, $x, y$ coordinates are also expressed relative to the corresponding grid cell, and can also be further away. For the width ($w$) and height ($h$) of human parts, we do not rely on YOLO's anchor system, since the selected anchors are not optimized for these targets. Instead, we predict values within the interval $[0-1]$, acting as scaling factors for the corresponding person's $w$ and $h$ predictions. The following equations compare how YOLO traditionally transforms detection outputs (Equation~\ref{eqn:yolo-inference}) versus YOLO-BKP's method for human parts (Equation~\ref{eqn:yolo-bkp-inference}). In a nutshell, there is no bounding interval for $x, y$, and $w, h$ scale the corresponding person detection.


\noindent\begin{minipage}{.55\linewidth}
\begin{equation}
\label{eqn:yolo-inference}
\begin{aligned}
  & x_b = 2\sigma(x) - 0.5 + grid_{i,j} \\
  & y_b = 2\sigma(y) - 0.5 + grid_{i,j} \\
  & w_b = \sigma(w)^2 * anchor_{a} \\
  & h_b = \sigma(h)^2 * anchor_{a} \\
  & conf_b = \sigma(conf)
\end{aligned}
\end{equation}
\end{minipage}%
\begin{minipage}{.45\linewidth}
\begin{equation}
\label{eqn:yolo-bkp-inference}
\begin{aligned}
  & x_p' = x_p + grid_{i,j} \\
  & y_p' = y_p + grid_{i,j} \\
  & w_p' = \sigma(w_p)^2 * w_b \\
  & h_p' = \sigma(h_p)^2 * h_b \\
  & conf_p' = \sigma(conf_p).
\end{aligned}
\end{equation}
\end{minipage}

\vspace{3mm}


To optimize the part detection head, we propose two additional loss components. Let us first see the original loss components for detection: an IoU-based loss to optimize the bounding box coordinates, and two binary cross-entropy components, one for the confidence estimate and another for class prediction. The following is a mathematical formulation for a ground truth matched with scale level $l$, grid cell $i, j$, and an anchor $a$. Consider $CIoU$ the Complete IoU proposed in Zheng et al.~\cite{zheng2021enhancing}.
\begin{equation}
\label{eq:lbox}
\begin{aligned}
    & IoU = CIoU(box_{gt}^{l,i,j,a}, box_{pred}^{l,i,j,a}) \\ 
    & \mathcal{L}_{box} = 1 - IoU  \\
    & \mathcal{L}_{conf} = BCE(conf_b, IoU)
\end{aligned}
\end{equation}


For pose estimation, there are two components: a distance metric for the $x, y$ coordinates and the confidence loss for visibility $v$. For the distance, \cite{maji2022yolo} proposed an adaptation of the Object-Keypoint Similarity (OKS) metric, commonly utilized to evaluate pose estimation. Both components are presented in Equation~\ref{eq:loss_kpt} for a set of keypoints $k$.
\begin{equation}
\label{eq:loss_kpt}
\begin{aligned}
      & OKS_k = exp(\frac{\displaystyle -\sqrt{(x_{k, gt}^{l,i,j,a} - x_{k, pred}^{l,i,j,a})^2 + (y_{k, gt}^{l,i,j,a} - y_{k, pred}^{l,i,j,a})^2}}{\displaystyle 2(area_{gt}^{l,i,j,a})^2 \omega_k^2}) \\
      & \mathcal{L}_{kpts} =  \frac{1}{\eta} \sum_{k=1}^{n} v_{k, gt} (1-OKS_k)  \\
      & \mathcal{L}_{kconf} = BCE(v_{pred}^{kpts} ,v_{gt}^{kpts})
\end{aligned}
\end{equation}


Consider $area_{gt}^{l,i,j,a}$ the area from the target bounding box of the corresponding person, $\omega$ a set of fixed weights for each keypoint. $\eta$ is the number of visible keypoints and $n$ the total number of keypoints. Consider $v_{pred}^{kpts}$ and $v_{gt}^{kpts}$ the visibility vector for all $n$ keypoints. In simple terms, confidence approximates the visibility flag and the coordinates are optimized according to a normalized distance metric.

We propose mixing both rationales, since our targets are multiple bounding boxes structured in a single vector. 
We use both the distance metric proposed in \cite{maji2022yolo}, and the IoU-based loss. 
The mathematical formulation of our proposed part box loss $\mathcal{L}_{pbox}$ is presented in Equation~\ref{eq:lpbox}. 
\begin{equation}
\label{eq:lpbox}
    \mathcal{L}_{pbox} = \frac{1}{\eta} \sum_{p=1}^{n} v_{p, gt} ( (1-IoU_p) + (1-OKS_p) )
\end{equation}

Consider $v_{p, gt}$ a binary visibility flag for the ground truth part $p$. While $IoU_p$ is calculated as defined in Equation~\ref{eq:lbox}, $OKS_p$ has particularities. First, it is calculated with a separate set of fixed weights $\gamma$ to define error tolerances for each part. Secondly, it is only performed over the $x, y$ center coordinates from the prediction and target boxes.

Our part confidence loss approximates the $IoU$ between target and prediction. Equation~\ref{eq:lpconf} presents this formulation, with $conf_{pred}^{parts}$ as the vector of predicted confidences, $v_{gt}^{parts}$ the set of visibilities from all targets, and $IoU^{parts}$ the $IoU$ calculated for all parts.
\begin{equation}
\label{eq:lpconf}
    \mathcal{L}_{pconf} = BCE(conf_{pred}^{parts}, v_{gt}^{parts} IoU^{parts})
\end{equation}

The total loss is the sum of components weighted by their respective $\lambda$. We retain $\lambda$ values established by Maji et al.~\cite{maji2022yolo}, and adopt the same $\lambda$ originally associated to $\mathcal{L}_{box}$ and $\mathcal{L}_{conf}$ for our proposed components, respectively $\mathcal{L}_{pbox}$ and $\mathcal{L}_{pconf}$.

\section{Experiments}
\label{sec:experiments}

This section evaluates our Human-Centric Perception (HCP) pipelines across multiple datasets. Section~\ref{sec:results-datasets} describes the datasets, including COCO for benchmarking (Section~\ref{sec:datasets-coco}) and RCPD for CSAI analysis (Section~\ref{sec:datasets-rcpd}). Section~\ref{sec:results-metrics} details the evaluation metrics. Section~\ref{sec:results-stats} evaluates our pipeline, while Section~\ref{sec:results-hcp-rcpd} presents insights from real CSAI.

\subsection{Datasets}
\label{sec:results-datasets} 

\subsubsection{COCO}
\label{sec:datasets-coco}

\begin{figure}[b]
    \centering
    \includegraphics[width=\linewidth]{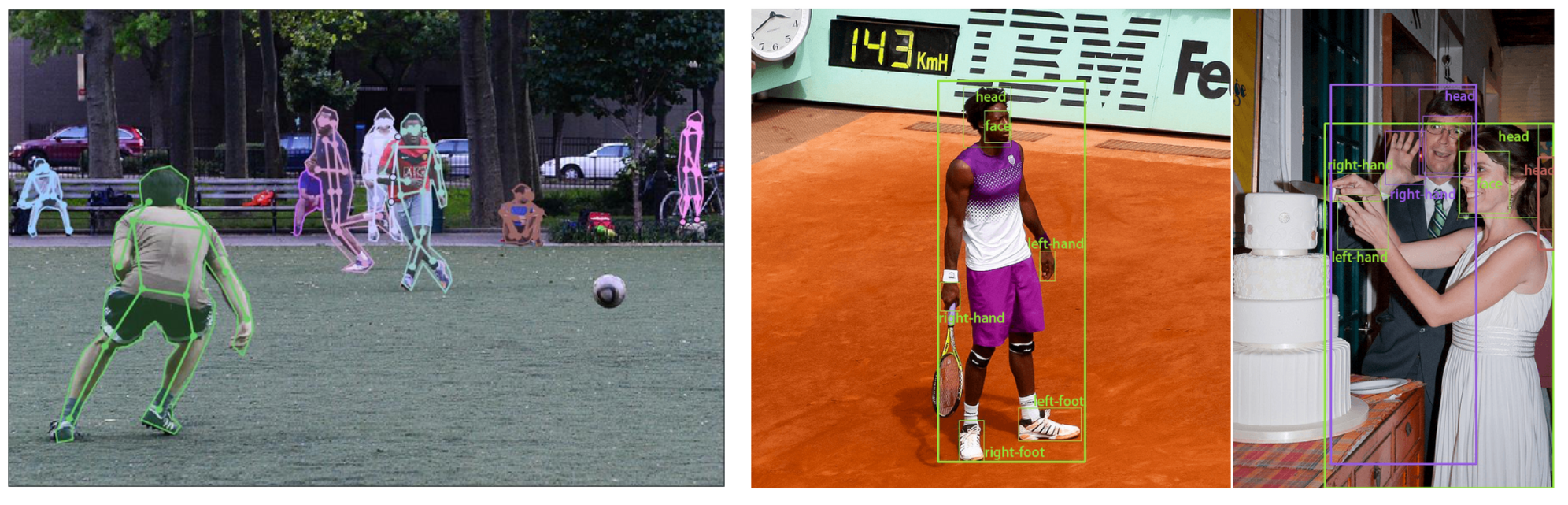}
    \caption[Samples with annotations from COCO-Keypoints and COCO-HumanParts.]{Samples with annotations from COCO-Keypoints (left) and COCO-HumanParts (right). }
    \label{fig:coco_samples}
\end{figure}

The Common Objects in Context (COCO) dataset~\cite{lin2014microsoft} provides annotations for all components required in our study. Specifically, the COCO-Keypoints includes person bounding boxes with associated keypoint coordinates. Additionally, \cite{yang2020hier} introduces COCO-HumanParts, incorporating hierarchical annotations for body parts, including head, face, hands, and feet. Figure~\ref{fig:coco_samples} shows samples from both datasets. 

From the \(\sim \)123k images on COCO2017, both datasets provide annotations for \(\sim \)66k images with at least one person. However, while both datasets annotate \(\sim \)270k person boxes, the number of annotated inner components varies drastically. Keypoint information is provided for \(\sim \)156k individuals, while \(\sim \)248k have at least one part box. Dividing person instances into the size categories adopted by Yang et al.~\cite{yang2020hier} reveals that small and tiny person boxes do not have keypoint information.




We combine labels from both data sources, producing a single Body-Keypoint-Part database similar to our own. Since the datasets were annotated separately, we match person instances across them using the provided person bounding boxes.

\subsubsection{RCPD}
\label{sec:datasets-rcpd}

RCPD~\cite{macedo2018benchmark} annotates person bounding boxes hierarchically associated with body parts (face, breast, genital, buttocks). Each bounding box in RCPD is accompanied by three class annotations: nudity level (none, semi-nude, nude, sexually active), binary gender expression, and perceived age. 

Our analysis is structured around two dimensions: (1) binary age (children aged $\leq13$, as per RCPD), and (2) a classification of CSAI, adult pornography, and other images. We define indecency by the presence of at least one nude or sexually active body part, and CSAI is distinguished from pornography by the presence of a child.


Table~\ref{tab:rcpd-label-count} dissects the labels from RCPD. Within the $837$~CSA images, there are \(\sim \)$960$ children and \(\sim \)$90$ older people. The vast majority of older people are perceived as male, and \(\sim \)$60\%$ of children perceived as female. The dataset has $508$ images without people. Thus, the remaining $793$ were classified as ``porn'' or ``other'' according to nudity levels, amounting respectively to $284$ and $509$ samples. The majority of people in adult pornography are perceived as female. Non-explicit images are balanced by gender expression.


\begin{table}[t]
\centering
\setlength\tabcolsep{0.5pt}
\begin{tabular*}{\linewidth}{@{\extracolsep{\fill}} lcccc|cc|cccc }
\toprule
Class  & CSAI & CSAI & CSAI & CSAI 
       & Porn & Porn 
       & Other & Other & Other & Other \\
Age    & $>$13 & $>$13 & $\le$13 & $\le$13 
       & $>$13 & $>$13 
       & $>$13 & $>$13 & $\le$13 & $\le$13   \\
Gender & M & F & M & F
       & M & F
       & M & F & M & F \\
\midrule
Person & 79 & 8 & 382 & 581 
       & 82 & 294 
       & 76 & 74 & 226 & 210 \\ \midrule
Face   & 6 & 6 & 324 & 451 
       & 75 & 280 
       & 76 & 74  & 226 & 209  \\ 
Breast & 6 & 4 & 120 & 193 
       & 26 & 290 
       & 17 & 30  & 52 & 17 \\ 
Genital & 75 & 5 & 264 & 330 
        & 49 & 165 
        &  0 & 0 & 0  & 0  \\ 
Buttocks & 0 & 0 & 23 & 32 
         & 2 & 15 
         & 0 & 0 & 1  & 0  \\ 

\bottomrule
\end{tabular*}
\caption{Number of detection labels in RCPD.}
\label{tab:rcpd-label-count}
\end{table}



Only $7.6\%$ of males aged $>13$ depicted in CSAI have their face or chest visible, whereas over $94\%$ display their genitals. The absence of visible faces and chests suggests a pattern of partial body visibility, consistent with statistics from SOD where hip labels are also more predominant (Table~\ref{tab:num-annot}). Among older females, $6$ out of $8$ have visible faces. Furthermore, regardless of gender, \(\sim \)$80\%$ of the children in CSAI have visible face. Thus, methods relying on facial features to estimate age would miss a substantial amount of children.

\subsection{Metrics}
\label{sec:results-metrics}

Our approach addresses three tasks, each with its own set of metrics. For \textbf{pose estimation}, we employ the OKS metric, as defined in the COCO evaluation API. To compute Average Precision (AP) and Average Recall (AR), true positives must exceed an OKS threshold. To evaluate \textbf{detection} for each target --- person and human parts --- we rely on the standard Intersection over Union (IoU)-based AP and AR. 
Lastly, to evaluate the \textbf{association} of person and human parts, works often resort to the COCO-HumanParts association metric, called APSub~\cite{yang2020hier}. However, this is an opaque metric, aggregating the evaluation of both the detection and the association as an average over all parts, lacking straightforward interpretation. Thus, we adopt the metrics proposed in \text{Narasimhaswamy et al.~\cite{narasimhaswamy2022whose}}, namely (1) Conditional Accuracy (CA), i.e., the percentage of correctly associated body parts from the set of true positives and (2) Joint Average Precision (JAP), which extends the standard AP, limiting true positives to parts correctly associated. Association metrics consider a fixed IoU threshold of $0.5$.

    



\subsection{Results}
\label{sec:results-stats} 
This section is divided into three parts. First, we benchmark our pipeline with a combination of well-established datasets, COCO-Keypoints and COCO-HumanParts (Section~\ref{sec:results-coco}). Next, we present experiments on our custom BKP-Dataset, firstly testing models previously trained on the combined COCO dataset (Section~\ref{sec:results-bkpd}), and later focusing on the novel detection targets by fine-tuning on BKPD (Section~\ref{sec:results-bkpd-tune}).


All experiments use the YOLOv7-Pose released publicly\footnote{\url{https://github.com/WongKinYiu/yolov7/tree/pose}}. For BKP-Association, the architecture remains unchanged, while YOLO-BKP incorporates a human parts detection head and corresponding loss components. Backbone weights are initialized and frozen using a COCO-Keypoints pre-trained model from the same code release.

\subsubsection{COCO}
\label{sec:results-coco}
Similar to other methods evaluated on \text{COCO-HumanParts}~\cite{yang2020hier, zhou2024bpjdet}, we adopt an input resolution of 1536$\times$1536 to capture part boxes of scale small and tiny. Training is conducted for 300 epochs using the SGD optimizer. We use YOLO-Pose's original hyperparameters and data augmentation. The only exception are the anchors, which are optimized to COCO-HumanParts by YOLO's autoanchor tool. For the non-maximum suppression of BKP-Association, we use the hyperparameters from \text{Zhou et al.~\cite{zhou2024bpjdet}}, $\tau_{conf}^b = 0.05 $, $\tau_{iou}^b = 0.6$, $\tau_{conf}^p = 0.1$, and $\tau_{iou}^p=0.3$. BKP-Association also requires defining the set of keypoints for the association stage of detected parts. For COCO, we use all five head keypoints for the head category, eyes and nose keypoints for the face, and associate each wrist and ankle keypoint with the corresponding hand and foot.



The first task to be evaluated is \textbf{pose estimation}. We experimented on the validation set of COCO2017, using COCO's official evaluation API. Table~\ref{tab:results-pose} benchmarks our methods. We only consider bottom-up approaches, i.e., predicting all individuals' poses directly from the entire image. Note that we work with the largest input resolution to account for human parts detection. However, since COCO-Keypoints is only annotated for medium and large person instances, metrics do not reflect performance for scales small and tiny. 

\begin{table}[tb]
    \centering
\setlength\tabcolsep{0.5pt}
\begin{tabular*}{\linewidth}{@{\extracolsep{\fill}} lllcccc }
    \toprule
    \textbf{Method} & \textbf{Backbone} & \textbf{Size} & \textbf{AP} & \textbf{AP$_{50}$} & \textbf{AR} & \textbf{AR$_{50}$} \\  \midrule
    Associate Embedding~\cite{newell2017associative} & ResNet50 & 512 & 46.6 & 74.2 & 56.6 & 81.0  \\ 
    Associate Embedding~\cite{newell2017associative} & ResNet101 & 512 & 55.4 & 80.7 & 62.2 & 84.1  \\ 
    Associate Embedding~\cite{newell2017associative} & ResNet152 & 512 & 59.5 & 82.9 & 65.1 & 85.6  \\ 
    Associate Embedding~\cite{newell2017associative} & Hourglass & 512 & 61.3 & 83.3 & 65.9 & 85.0  \\ 
    HigherHRNet~\cite{cheng2020higherhrnet} & HRNet-W32 & 512 & 67.7 & 87.0 & 72.9 & 89.0  \\ 
    HigherHRNet~\cite{cheng2020higherhrnet} & HRNet-W48 & 512 & 68.6 & 87.3 & 73.1 & 89.2  \\ 
    HigherHRNet~\cite{cheng2020higherhrnet} & HRNet-W32 & 640 & 68.5 & 87.1 & -- & --  \\ 
    HigherHRNet~\cite{cheng2020higherhrnet} & HRNet-W48 & 640 & 69.9 & 87.2 & -- & --  \\ 
    DEKR~\cite{geng2021bottom} & HRNet-W32 & 512 & 67.3 & 87.9 & 72.4 & 89.8  \\ 
    DEKR~\cite{geng2021bottom} & HRNet-W48 & 640 & 70.0 & 89.4 & 75.4 & 91.4  \\ 
    VitPose-B~\cite{xu2023vitpose} & VIT-B & 512 & 68.5 & 87.3 & 72.9 & 89.2 \\
    VitPose-L~\cite{xu2023vitpose} & VIT-L & 512 & \textbf{70.1} & 88.3 & 74.5 & 90.2 \\    
    YOLOv5s6-pose~\cite{maji2022yolo} & YOLOv5s6 & 960  & 63.8 & 87.6 & 70.4 & --  \\ 
    YOLOv5m6-pose~\cite{maji2022yolo} & YOLOv5m6 & 960  & 67.4 & 89.1 & 73.9 & --  \\ 
    YOLOv5l6-pose~\cite{maji2022yolo} & YOLOv5l6 & 960  & 69.4 & \textbf{90.2} & \textbf{75.9} & --  \\ 
    YOLOv7-w6-pose~\cite{maji2022yolo} & YOLOv7-W6 & 960  & 70.0 & 88.6 & 75.6 & \textbf{91.9}  \\ \midrule
    BKP-Association* & YOLOv7-W6 & 1536 & 69.5 & 87.6 & 75.7 & \textbf{91.9} \\ 
    YOLO-BKP & YOLOv7-W6 & 1536 & 68.0 & 87.4 & 74.4 & 91.5 \\
    \bottomrule
\end{tabular*}
\caption{Pose estimation results for bottom-up methods on COCO2017 val set. SOTA results are presented as reported for single-scale testing. Our methods are trained on the joint COCO-Keypoints-HumanParts dataset.\\ \small{*BKP-Association's contribution is the association algorithm, which does not impact the pose estimation task.}}
\label{tab:results-pose}
\end{table}

\begin{figure}[t]
    \centering
    \includegraphics[width=\linewidth]{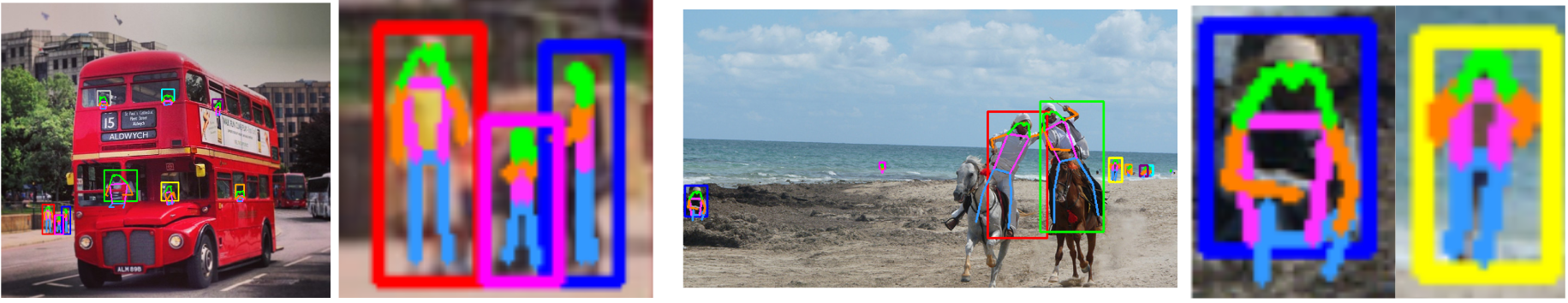}
    \caption{Qualitative results for pose estimation for small and tiny persons. Original images and zoom-ins side-by-side.}
    \label{fig:skeleton-tiny}
\end{figure}

Sexually explicit domains primarily depict humans at larger scales. For COCO, we need to ensure keypoints are reliable at smaller scales. Figure~\ref{fig:skeleton-tiny} demonstrates the model's ability to approximate keypoints at reduced scales. Future research could explore whether joint training with human parts, with annotations at smaller scales, enhances this capability.

Moving on to the task of \textbf{detection}, Table~\ref{tab:results-coco-humanparts} presents the results reported in the literature compared to our propositions. Up to this point, the association algorithm proposed for BKP-Association has no influence on the results, since the detection outputs will feed the algorithm. Thus, BKP-Association performs similarly to the anchor-based versions of BPJDet (S, M, and L), since they all rely on similar versions of YOLO.

      






\begin{table*}[tb]
    \centering
    \setlength\tabcolsep{0.5pt}
    \begin{tabular*}{0.8\linewidth}{@{\extracolsep{\fill}} lccccccc|ccccccc }
    \toprule
    \multirow{2}{*}{\textbf{Method}} & \multicolumn{7}{c|}{\textbf{All categories}} & \multicolumn{7}{c}{\textbf{Per category AP}} \\ 
                  & AP   & AP$_{50}$ & AP$_{75}$ & AP$_T$ & AP$_S$ & AP$_M$ & AP$_L$    
                  & person & head     & face   & r-hand & l-hand & r-foot & l-foot \\ \midrule
    Faster-C4-R50~\cite{ren2016faster}& 32.0   & 55.5   & 32.3 & 9.9 & 38.5  & 54.9   & 52.4 
                  & 50.5   & 47.5     & 35.5   & 27.2   & 24.9  & 19.2 & 19.3 \\
    RetinaNet-R50~\cite{ross2017focal}& 32.2   & 54.7     & 33.3  & 10.0 & 39.8 & 54.5   & 53.8                              & 49.7   & 47.1     & 33.7   & 28.7   & 26.7  & 19.7 & 20.2 \\
    Faster-FPN-R50~\cite{he2017mask}& 34.8   & 60.0     & 35.4 & 14.0 & 41.8   & 55.4   & 52.2                              & 51.4   & 48.7     & 36.7   & 31.7   & 29.7  & 22.4 & 22.9 \\
    Hier-R50~\cite{yang2020hier} & 36.8   & 65.7     & 36.2 & 19.9 & 41.9   & 53.9   & 47.5
                  & 53.2   & 50.9     & 41.5   & 31.3   & 29.3  & 25.5 & 26.1
    \\ \midrule
    Faster-FPN-R101~\cite{he2017mask}& 36.0  & 62.1     & 36.5 & 14.8 & 43.0  & 57.2   & 54.8
                  &  52.6  & 49.3     & 36.9   & 33.3   & 30.3  & 24.3 & 24.4 \\ 
    Hier-R101~\cite{yang2020hier} & 37.2   & 65.9     & 36.7 & 19.4 & 42.3  & 55.1   & 50.3
                  & 54.0   & 50.4     & 41.6   & 31.6   & 30.1  & 26.0 & 26.6 \\ 
    Faster-FPN-X101~\cite{he2017mask}& 36.7  & 62.8     & 37.4 & 15.4 & 44.0  & 57.4   & 55.3
                   & 53.6  & 49.7     & 37.3   & 33.8   & 32.2  & 25.0 & 25.1 \\
    Hier-X101~\cite{yang2020hier} &  38.8  & 68.1     & 38.5 & 20.6 & 44.5 & 56.6   & 52.3
                  &  55.4  & 52.3     & 43.2   & 33.5   & 32.0  & 27.4 & 27.9
    \\ \midrule
    BPJDet-S~\cite{zhou2024bpjdet} & 38.9   & 65.5     & 39.4  & 19.1 & 46.3     & 59.1   & 49.7
                  & 56.3   & 53.5     & 41.9   & 34.7   & 33.7  & 25.5 & 26.5 \\
    BPJDet-M~\cite{zhou2024bpjdet} & 42.0   & 68.9     & 43.2 & 21.8 & 49.8      & 62.3   & 54.6
                  & 59.8   & 55.7     & 44.7   & 38.7   & 37.6  & 28.6 & 29.2 \\
    BPJDet-L~\cite{zhou2024bpjdet} & 43.6   & 70.6     & 45.1  & 23.3 & 51.3 & 63.8   & 61.8
                  & 61.3   & 56.6     & 46.3   & 40.4   & 39.6  & 30.2 & 30.9 \\ \midrule
    BPJDet-L5u~\cite{zhou2024bpjdet} & 44.0   & 71.9     & 44.8 & -- & --  & 65.7   & 66.5
                  & 63.7   & 55.4     & 47.0   & 40.8   & 39.3  & 30.5 & 31.5 \\
    BPJDet-L8u~\cite{zhou2024bpjdet} & 43.5   & 71.7     & 44.2 & -- & --  & 65.4   & 63.1
                  & 60.6   & 55.3     & 47.1   & 40.5   & 39.4  & 30.4 & 31.2 
    \\ \midrule
    BKP-Assoc. & 42.1  &  70.5 & 42.7 & 21.4 & 50.0 & 62.2   & 56.7 
                  & 60.5   & 55.3     & 46.0   &  37.9  & 36.6  & 28.7 & 29.3 \\ 
    YOLO-BKP & 39.4  &  69.9 & 38.2 & 21.5 & 46.0 & 55.5   & 48.6 
                  & 61.6   & 54.5     & 46.3   & 30.9   & 29.9  & 26.1 & 26.4 
    \\ \bottomrule
    \end{tabular*}
    \caption[Detection results on COCO-HumanParts val set.]{Detection results on COCO-HumanParts val set. SOTA results are presented as reported for single-scale testing. Our methods are trained on the joint COCO-Keypoints-HumanParts dataset.}
    \label{tab:results-coco-humanparts}
\end{table*}

For YOLO-BKP, upon looking at the AP per category, it is the best performing for person detection, competitive on the categories head and face, with lower AP for hands and feet. This may be due to the challenge of detecting smaller body parts with features from the same scale as the person. Furthermore, focusing on the summarized $AP_{50}$, YOLO-BKP's performance aligns with the state of the art indicating competitive, although coarse, detection capabilities.

Lastly, we evaluate \textbf{association} of bodies and parts. As we employ metrics better suited for detailed performance analysis, no results for Joint AP and Conditional Accuracy are reported by previous work leveraging COCO-HumanParts. We selected two representative models: Hier-R50~\cite{yang2020hier}, a baseline proposed along with COCO-HumanParts, and BPJDet-L~\cite{zhou2024bpjdet}, the top-performing anchor-based variant of BPJDet. Both offer publicly available trained models and source codes.

\begin{table}[tb]  
    \centering
    \setlength\tabcolsep{0.5pt}
    \begin{tabular*}{\linewidth}{@{\extracolsep{\fill}} lcccc|cccc}
    \toprule
    \textbf{Method}            & \textbf{Part} &  \textbf{AP$_{50}$}  & \textbf{JAP} & \phantom{a}\textbf{CA}\phantom{a} 
                      & \textbf{Part} &  \textbf{AP$_{50}$}  & \textbf{JAP} & \textbf{CA} \\ \midrule
                                
    Hier-R50~\cite{yang2020hier}& H     & 78.0  & 73.4  & 93.1
                      & F     & 68.7  & 64.8  & 93.5  \\ 
    BPJDet-L~\cite{zhou2024bpjdet}& H     & 82.5  & 76.8  & 95.6
                      & F     & 71.2  & 64.9  & 94.2  \\ 
    BKP-Assoc.&  H & 81.6 & 74.8 & 94.5
                      & F     & 72.1 & 65.1 & 93.7  \\ 
    YOLO-BKP     & H     & \textbf{82.9}  & \textbf{79.5}  & \textbf{95.7}
                      & F     & \textbf{76.9}  & \textbf{73.8}  & \textbf{95.8}  \\ \midrule

    Hier-R50~\cite{yang2020hier}& LH   & 49.6  & 47.6  & 94.0
                      & RH   & 52.3  & 50.1  & 93.8  \\ 
    BPJDet-L~\cite{zhou2024bpjdet}& LH   & \textbf{65.5}  & \textbf{61.4}  & 96.0
                      & RH   & \textbf{65.0}  & \textbf{60.6}  & 95.8  \\ 
    BKP-Assoc. &  LH & 62.1 & 56.6 & 93.7
                      & RH   & 62.9 & 57.4 & 93.9 \\ 
    YOLO-BKP     & LH   & 61.5  & 60.0  & \textbf{97.6}
                      & RH   & 61.3  & 59.9  & \textbf{97.7}  \\ \midrule

    Hier-R50~\cite{yang2020hier}& LF   & 51.4  & 50.5  & 97.6
                      & RF   & 49.5  & 48.8  & 98.3  \\ 
    BPJDet-L~\cite{zhou2024bpjdet}& LF   & \textbf{66.6}  & \textbf{65.8}  & 99.1
                      & RF   & \textbf{65.6}  & \textbf{64.6}  & 99.0  \\ 
    
    BKP-Assoc. & LF & 64.4 & 62.4 & 97.5
                      & RF   & 62.5 & 60.5 & 97.7  \\ 
    YOLO-BKP & LF   & 66.0  & 65.5  & \textbf{99.4}
                      & RF   & 64.5  & 64.1  & \textbf{99.6}  \\ \bottomrule
    \end{tabular*}
    \caption{Association results on COCO-HumanParts val set. H: Head, F: Face, LH: Left Hand, RH: Right Hand, LF: Left Foot, RF: Right Foot.}
    \label{tab:results-association}
\end{table}

Table~\ref{tab:results-association} summarizes association results. 
YOLO-BKP performs better, yielding the highest CA and the smallest AP-to-JointAP decline. It significantly outperforms head and face detection competitors and matches leading methods in JointAP for other targets. While detections are coarser, YOLO-BKP more effectively constructs a complete human representation. For BKP-Association, leveraging keypoints for the association results in slightly lower conditional accuracy, as keypoints are not optimized to align with part boxes. However, it remains competitive in detection performance.

\begin{figure}[t]
    \centering
    \includegraphics[width=\linewidth]{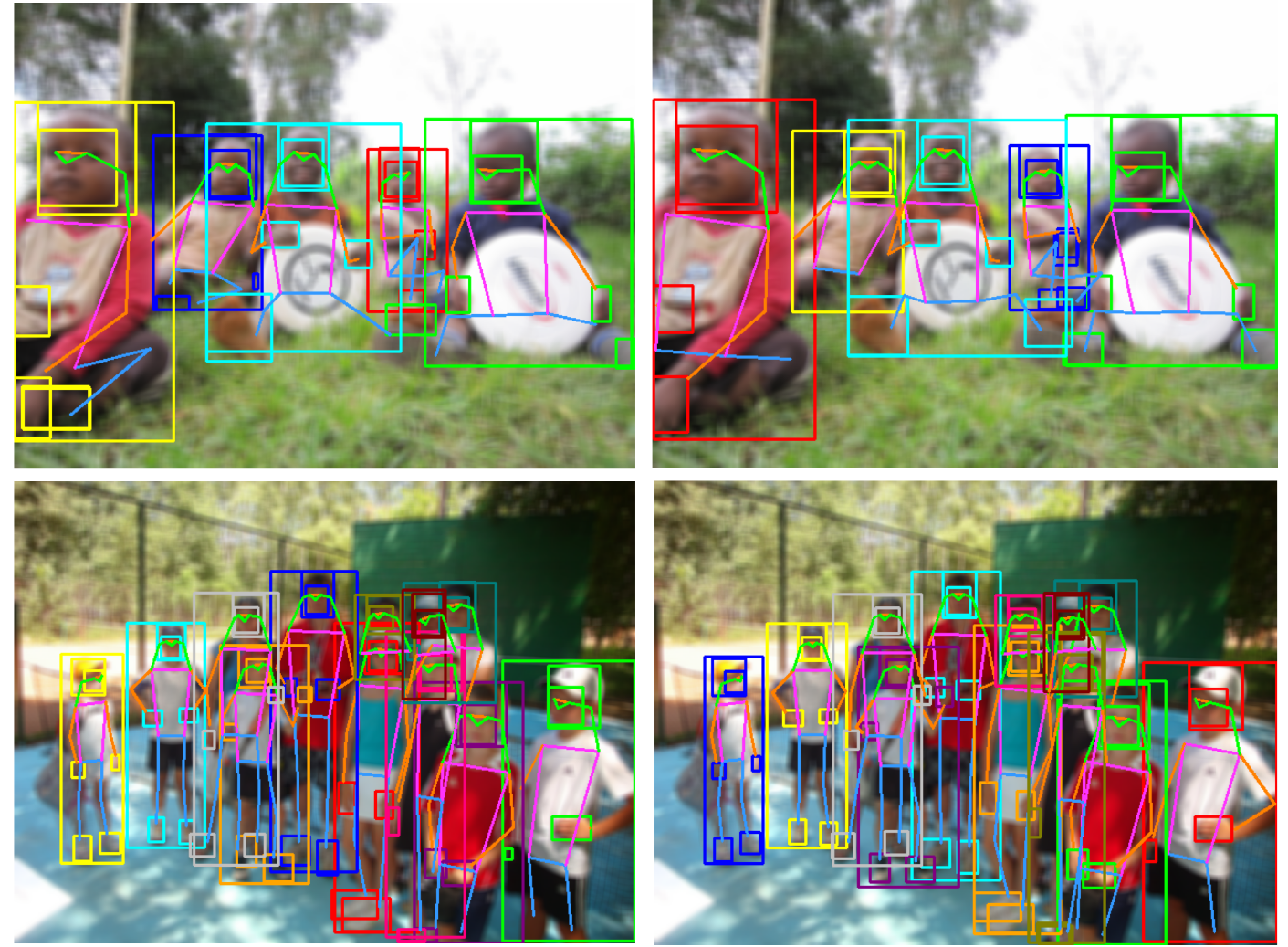}
    \caption{Qualitative results on COCO2017 val set. Left: BKP-Association, right: YOLO-BKP. Bounding box colors indicate association to the same individual.}
    \label{fig:qualitative-results}
\end{figure}

Qualitative results are presented in Figure~\ref{fig:qualitative-results}. BKP-Association provides precise detections, with tighter bounding boxes. We observed more instances of incorrect associations, as illustrated in the first row of Figure~\ref{fig:qualitative-results}, where the left foot of the central child (in green) is mistakenly associated to the child in yellow. YOLO-BKP, in contrast, infers part locations based on the overall body structure. It produces coarser detections, extending beyond the part boundaries, but it successfully identifies and associates parts that BKP-Association misses.


\subsubsection{BKPD -- Models Trained on COCO}
\label{sec:results-bkpd}
In this setting, results are presented only for targets common across COCO and BKPD. We divide results into the subset derived from Open Images (referred to as BKPD-OI), and the subset sourced from SOD (BKPD-SOD), the latter being a more pronounced domain shift compared to the training data. 

For pose estimation, Table~\ref{tab:results-pose-bkpd} compares our results with the best-performing methods, showing competitive performance. BKPD-SOD presents a greater challenge, as sexually explicit images may depict complex poses and intricate interactions. 
Conversely, BKPD-OI is an easier target, likely due to our selection bias of samples with fewer and salient people.

\begin{table}[tb]
    \centering
\setlength\tabcolsep{0.5pt}
\begin{tabular*}{\linewidth}{@{\extracolsep{\fill}} lllcccc|cccc }
    \toprule
     \multirow{2}{*}{\textbf{Method}}   & & & 
                   \multicolumn{4}{c|}{\textbf{BKPD-OI}} &
                   \multicolumn{4}{c}{\textbf{BKPD-SOD}} \\ 
                   & & & AP & AP$_{50}$ & AR & AR$_{50}$
                   & AP & AP$_{50}$ & AR & AR$_{50}$
                   \\  \midrule
    DEKR-w32~\cite{geng2021bottom} & & & 85.3 & 93.7 & 90.7 & 97.9  
                                           & 67.8 & 83.1 & 72.7 & 87.7 \\
    DEKR-w48~\cite{geng2021bottom} & & & 83.5 & 91.8 & 91.4 & 98.3  
                                           & 67.7 & 82.9 & 73.7 & 89.4 \\
    YOLOv7-w6-pose~\cite{maji2022yolo}& & & 91.6 & 96.6 & 93.9 & 98.0  
                                               & 68.9 & 85.1 & 75.0 & 87.7\\ 
    BKP-Assoc.  & & & 88.9 & 95.6 & 91.9 & 97.6 
                                & 67.6 & 84.3 & 72.9 & 86.6 \\      
    YOLO-BKP  & & & 88.5 & 96.0 & 91.9 & 98.0 
                                & 66.0 & 83.9 & 73.5 & 87.6 \\ \bottomrule
\end{tabular*}
\caption[Pose estimation results on BKPD for bottom-up methods trained on COCO-Keypoints.]{Pose estimation results on BKPD for bottom-up methods trained on COCO-Keypoints, except our methods, which are trained on the joint COCO-Keypoints-HumanParts.}
\label{tab:results-pose-bkpd}
\end{table}

For the remaining experiments, we adopted the input size $960$px as BKPD is a human-centric dataset and preliminary experiments showed better performance at this scale. Since our methods and BPJDet are fully convolutional, they support flexible input sizes at test time. This does not apply to \text{Hier-R-CNN}, so we report its results at the original scale.

Table~\ref{tab:results-detection-bkpd} summarizes results for the detection task on BKPD. 
It strengthens our argument that YOLO-BKP matches the state of the art on coarser detections. Nevertheless, this approach falls behind on the stricter AP metrics, despite achieving competitive results for the person and head categories. BKP-Association approximates the results of \text{BPJDet-L}, as both methods share significant similarities in detection architecture and the association as a second stage.


\begin{table}[tb]
    \centering
    \footnotesize\setlength\tabcolsep{0.5pt}
    \begin{tabular*}{\linewidth}{@{\extracolsep{\fill}} lccccccc|cccc }
    \toprule
    \multirow{2}{*}{\textbf{Method}} & \multirow{2}{*}{\textbf{Size}} & \multicolumn{6}{c|}{\textbf{All categories}} & \multicolumn{4}{c}{\textbf{Per category AP}} \\ 
                  & & \phantom{a}AP\phantom{a}   & AP$_{50}$ & AP$_{75}$ &   AP$_S$ & AP$_M$ & AP$_L$    
                  & Person & Head     &  \phantom{a}LH\phantom{a} & RH  \\ \midrule
                  & \multicolumn{8}{c}{BKPD-OI} \\  \midrule
    Hier-R50~\cite{yang2020hier}& 1536 & 50.2   &  79.0 &  52.5  & 23.1   & 40.6   & 59.0
                  & 70.8   & 65.3     & 31.6   & 33.0   \\
        

    BPJDet-L~\cite{zhou2024bpjdet} & 960 & 61.9   &  86.6 & 66.5   & 23.4 & \textbf{48.7}  & 70.8 
                  & 83.0   & 72.9  & 44.5   & 47.3   \\
    BKP-Assoc.  & 960 & \textbf{62.7}  &  \textbf{88.2} & \textbf{67.2}  & \textbf{24.9}  & 48.5   & \textbf{71.8} 
                  & 84.3   & \textbf{73.4}     & \textbf{45.3}   & \textbf{47.8}  \\
                  
    YOLO-BKP  & 960 & 59.1  & 87.4 & 61.3 & 21.0 & 47.2 & 65.9 
                  & \textbf{85.8}   & 71.4   & 39.6  & 39.7 \\  \midrule
    
                  & \multicolumn{8}{c}{BKPD-SOD} \\  \midrule
    Hier-R50~\cite{yang2020hier}& 1536 & 51.7   &  77.0   & 58.6     & -- & 29.7   & 55.6
                  & 69.1   & 65.4     & 36.2   & 36.2  \\
    BPJDet-L~\cite{zhou2024bpjdet}& 960 & \textbf{63.8}   &  86.4  & 72.1   & --  & \textbf{43.0}   & \textbf{67.4} 
                  & \textbf{83.0}   & 72.4    & \textbf{49.8}   & \textbf{50.0}  \\
    BKP-Assoc.   & 960 & 63.5  &  \textbf{86.6}  & \textbf{72.2}   &  --  & 42.6  & 66.9 
                  & 82.2   & \textbf{72.7}     & 49.4   & 49.6 \\ 
    YOLO-BKP & 960 & 60.5  &  86.4  & 65.5  & --  & 39.2  & 63.5 
                  & 82.7   & 70.0   & 44.0  & 45.4 
                  
    \\ \bottomrule
    \end{tabular*}
    \caption[Detection results on BKPD for models trained on COCO-HumanParts.]{Detection results on BKPD for models trained on COCO-HumanParts, except our methods, which are trained on the joint COCO-Keypoints-HumanParts.}
    \label{tab:results-detection-bkpd}
\end{table}

Table~\ref{tab:results-association-bkpd} presents the association results on BKPD, highlighting the advantages of YOLO-BKP. It yields the best conditional accuracy among tested approaches and consistently outperforms existing approaches for the head category. Furthermore, YOLO-BKP also achieved superior performance for both hand classes, further widening the performance gap for JointAP. It suggests that for a human-centered database, our method might be the best choice for a complete representation of the human body in a context that allows coarser detections.


BKP-Association's performance is comparable to that of BPJDet. The key distinction lies in the association step, in which BKP-Association avoids training redundant inferences for human parts, using keypoints as reference for association. Our findings indicate that pose information can effectively support the association of human body parts.


\begin{table}[tb]  
    \centering
    \footnotesize\setlength\tabcolsep{0.5pt}
    \begin{tabular*}{\linewidth}{@{\extracolsep{\fill}} lcccc|ccc|ccc}
    \toprule
    \multirow{2}{*}{\textbf{Method}} & 
    \multirow{2}{*}{\textbf{Size\phantom{a}}}
                      &  \multicolumn{3}{c|}{\textbf{Head}}
                      &  \multicolumn{3}{c|}{\textbf{L-Hand}}
                      &  \multicolumn{3}{c}{\textbf{R-Hand}} \\
                      & &  AP$_{50}$  & JAP & \phantom{a}CA\phantom{a} 
                      &  AP$_{50}$  & JAP & \phantom{a}\phantom{a}CA\phantom{a}
                      &  AP$_{50}$  & JAP & CA \\ \midrule
                      & \multicolumn{9}{c}{\textbf{BKPD-OI}} \\  \midrule
                                
    Hier-R50~\cite{yang2020hier}& 1536 & 93.5  & 88.8  & 95.5
                      & 66.5  & 62.5  & 93.7 
                      & 68.1  & 63.6  & 92.8 \\ 
    BPJDet-L~\cite{zhou2024bpjdet}& 960 & 95.5  & 91.5  & \textbf{97.4}
                      & 78.1  & 73.4  & 96.0 
                      & \textbf{79.2}  & 73.0  & 94.8 \\
    BKP-Assoc.  & 960 & 97.1  & 92.0  & 96.9
                      & 77.2  & 68.6  & 92.4 
                      & 78.5  & 70.6  & 92.7 \\ 
    YOLO-BKP    & 960 & \textbf{97.6}  & \textbf{95.1}  & \textbf{97.4}
                      & \textbf{78.4}  & \textbf{76.1}  & \textbf{96.6} 
                      & 77.6  & \textbf{75.0}  & \textbf{96.8} \\ \midrule
                      
                      & \multicolumn{9}{c}{\textbf{BKPD-SOD}} \\  \midrule
                                
    Hier-R50~\cite{yang2020hier}& 1536 & 89.0  & 79.9  & 89.5
                      & 65.6  & 61.5  & 92.8 
                      & 65.9  & 62.0  & 93.5 \\
    BPJDet-L~\cite{zhou2024bpjdet}& 960 & 94.8  & 89.8  & \textbf{95.9}
                      & \textbf{78.2}  & 73.9  & 96.3 
                      & 78.4  & 73.5  & 95.5 \\
    BKP-Assoc. & 960 & \textbf{95.4}  & 90.5  & 95.8
                      & 76.7  & 70.2  & 94.3 
                      & 78.0  & 72.4  & 95.4 \\                  
    YOLO-BKP  & 960  & 94.3  & \textbf{91.1}  & 95.8
                      & 77.7  & \textbf{75.0}  & \textbf{96.4} 
                      & \textbf{80.8}  & \textbf{78.1 } & \textbf{96.1} 
                                

                      \\ \bottomrule
    \end{tabular*}
    \caption[Association results on BKPD for models trained on COCO-HumanParts.]{Association results on BKPD for models trained on COCO-HumanParts, with the exception of our methods, which are trained on the joint COCO-Keypoints-HumanParts.}
    \label{tab:results-association-bkpd}
\end{table}

\subsubsection{Fine-tuning on BKPD}
\label{sec:results-bkpd-tune}

In this stage, we also freeze the Path Aggregation (PANet), training only the inference heads for 200 epochs with the same hyperparameters and data augmentations utilized in the previous training. We set the input resolution $960 \times 960$ as per YOLO-Pose~\cite{maji2022yolo}. For the association step of BKP-Association, we leverage all five head keypoints to associate the head, shoulder keypoints for the chest, hip keypoints for the hip, and each wrist coordinate for its corresponding hand.


Training and testing splits have equal proportions of samples from Open Images (OI) and SOD to assess model performance under consistent domain conditions. Cross-domain experiments evaluate robustness: models were trained on the Open Images subset of BKPD and tested on SOD, and vice versa. For comparability with the 50-50 split, cross-domain experiments used the entire source domain for training, and validation samples from the target domain for testing.

Table~\ref{tab:results-association-robustness} summarizes both the detection and association capabilities of our models on BKPD. Models trained on the complete BKP-Dataset showed the highest performance compared to cross-domain experiments. YOLO-BKP exhibits a greater ability to learn novel targets --- chest and hip --- when evaluated on Open Images, regardless of the training set. BKP-Association outperforms YOLO-BKP's precision for novel targets in SOD, provided the model is trained on the same domain. BKP-Association is better at retaining prior knowledge, likely due to its simpler detection head than YOLO-BKP’s more complex human part detection. This is reflected in its superior results for detecting head and hands.



\begin{table*}[t]  
    \centering
    \setlength\tabcolsep{0.5pt}
    \begin{tabular*}{\linewidth}{@{\extracolsep{\fill}} lcc|cccc|cccc|cccc|cccc|cccc}
    \toprule
    \multirow{2}{*}{\textbf{Method}} & \multirow{2}{*}{\textbf{Tune}} & \multirow{2}{*}{\textbf{Val}}
                      &  \multicolumn{4}{c|}{\textbf{Head}}
                      &  \multicolumn{4}{c|}{\textbf{Chest}}
                      &  \multicolumn{4}{c|}{\textbf{Hip}}
                      &  \multicolumn{4}{c|}{\textbf{L-Hand}}
                      &  \multicolumn{4}{c}{\textbf{R-Hand}}\\
                & &   &  AR$_{50}$  
                      &  AP$_{50}$  & JAP & \phantom{a}CA\phantom{a} 
                      &  AR$_{50}$  
                      &  AP$_{50}$  & JAP & \phantom{a}CA\phantom{a}
                      &  AR$_{50}$  
                      &  AP$_{50}$  & JAP & \phantom{a}CA\phantom{a}
                      &  AR$_{50}$  
                      &  AP$_{50}$  & JAP & \phantom{a}CA\phantom{a}
                      &  AR$_{50}$  
                      &  AP$_{50}$  & JAP & CA \\ \midrule

    BKP-Assoc. & All & OI  
               & 95.2 & 96.2  & 91.0  & 96.7
               & 62.6 & 65.1  & 62.1  & 97.5 
               & 65.5 & 64.8  & 62.5  & 97.9
               & \textbf{78.3} & \textbf{78.1}  & \textbf{68.8}  & 91.8
               & \textbf{77.3} & \textbf{74.9}  & \textbf{64.0}  & 92.2 \\  
               
    YOLO-BKP   & All & OI  
               & \textbf{97.}5 & \textbf{97.1}  & \textbf{95.4}  & \textbf{97.9}
               & \textbf{83.2} & \textbf{76.7}  & \textbf{75.1}  & \textbf{98.1}
               & \textbf{84.3} & \textbf{76.1}  & \textbf{75.7}  &  \textbf{99.6} 
               & 77.0 & 70.0  & 68.3  & \textbf{97.8}
               & 63.9 & 65.3  & 64.3  & \textbf{97.0} \\ \midrule

    BKP-Assoc. & SOD & OI  
               & 88.5 & 93.6  & 84.9  & 94.1
               & 34.9 & 23.6  & 21.6  & 94.6 
               & 44.6  & 39.5  & 38.5  & 98.2 
               & \textbf{72.0} & \textbf{71.7}  & \textbf{62.1}  & 91.8
               & \textbf{71.6} & \textbf{70.2}  & \textbf{58.4}  & 90.8 \\  
    YOLO-BKP   & SOD & OI  
               & \textbf{94.7} & \textbf{93.8} & \textbf{92.1} & \textbf{97.6}
               & \textbf{68.0} & \textbf{54.5} & \textbf{53.4} & \textbf{97.9}
               & \textbf{69.7} & \textbf{57.3} & \textbf{57.1} & \textbf{99.3} 
               & 71.1 & 63.4 & \textbf{62.1} & \textbf{96.7}
               & 65.8 & 58.0 & 56.7 & \textbf{97.1}
               \\ \midrule \midrule

    BKP-Assoc. & All & SOD  
               & \textbf{94.4} & \textbf{95.1}  & \textbf{86.7}  & \textbf{95.2}
               & 59.8 & \textbf{60.7}  & \textbf{58.0}  & 96.7 
               & 69.9 & \textbf{72.4}  & \textbf{69.5}  & 96.7
               & \textbf{76.9} & \textbf{76.9}  & \textbf{70.7}  & 93.9
               & \textbf{82.3} & \textbf{80.7}  & \textbf{75.5}  & 94.8 \\

    YOLO-BKP   & All & SOD 
               & 94.2 & 93.8  & 90.5  & 93.6
               & \textbf{60.3} & 51.3  & 50.7  & \textbf{98.8} 
               & \textbf{72.6} & 67.9  & 66.8 & \textbf{97.9} 
               & 75.6 & 75.0  & 64.9  & \textbf{91.2}
               & 75.6 & 79.7  & 67.8  & \textbf{90.0}
               \\ \midrule

    BKP-Assoc. & OI & SOD  
               & \textbf{93.8} & \textbf{94.7}  & 85.3  & \textbf{93.9}
               & 42.1 & 35.3  & 33.2  & 96.9
               & 39.2 & 33.8  & 31.6  & 96.2 
               & \textbf{76.6} & \textbf{76.7}  & \textbf{66.4}  & 92.0
               & 72.9 & \textbf{76.0}  & \textbf{67.4}  & 92.4 \\  
    YOLO-BKP   & OI & SOD  
               & 92.3 & 91.3 & \textbf{85.9} & 92.6
               & \textbf{47.6} & \textbf{36.7} & \textbf{36.6} & \textbf{99.9}
              & \textbf{45.4} & \textbf{39.3} & \textbf{39.3} & \textbf{100} 
              & 69.7 & 65.3 & 61.5 & \textbf{94.8}
               & 72.9 & 66.5 & 62.1 & \textbf{92.8}
               \\ \bottomrule
               
    \end{tabular*}
    \caption{Detection and association performance with various data splits from BKPD.}
    \label{tab:results-association-robustness}
\end{table*}

\begin{figure}[t]
    \centering
    \includegraphics[width=0.9\linewidth]{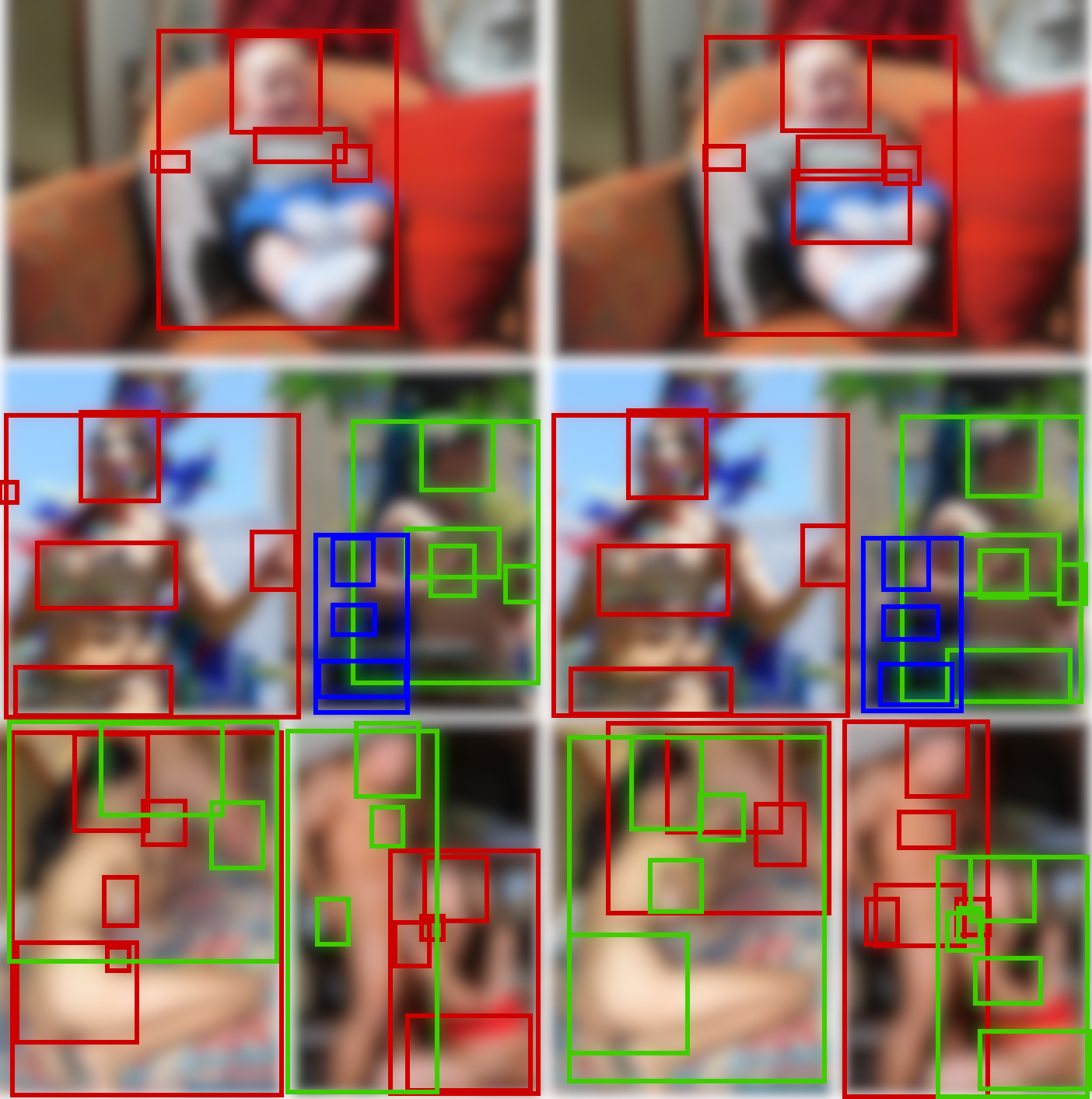}
    \caption{Example hierarchic detections from BKP-Association (left) and YOLO-BKP (right).}
    \label{fig:BKPD-detections}
\end{figure}

For a better understanding of the disparity between both of our approaches, we display in Figure~\ref{fig:BKPD-detections} example detections on samples from Open Images and SOD. They highlight that YOLO-BKP maintains its tendency for larger and less precise detections than BKP-Association. The primary observation regarding the lower performance of BKP-Association on non-explicit samples is its tendency to miss visible parts, as illustrated in the first and third images of Figure~\ref{fig:BKPD-detections}. We hypothesize that this behavior stems from its reliance solely on local features. In contrast, YOLO-BKP, which leverages features from the same scale as the person detection stream, has access to global information. This broader perspective likely provides critical contextual cues. Despite the performance disparity in the domain of sexually explicit images, YOLO-BKP is considered the most robust model in contexts where coarse detections are sufficient, particularly given YOLO-BKP's superior recall, a crucial concern for CSAI.

Lastly, Table~\ref{tab:results-association-robustness} shows overall lower detection and association performance for the chest and hip classes, particularly in adult pornography. This behavior is discussed in more depth in Section~\ref{sec:results-hcp-rcpd}, which examines how our methods tend to concentrate detections around the person’s torso, partially excluding body parts that extend beyond this region.

\subsection{Case Study: RCPD}
\label{sec:results-hcp-rcpd}
In this section, we apply our human-centric pipeline to RCPD~\cite{macedo2018benchmark}, a CSAI dataset. As detection targets are distinct between RCPD and BKPD we raise the proper considerations for quantitative assessment of each discussed target. 

\begin{figure}[t]
    \centering
    \includegraphics[width=0.75\linewidth]{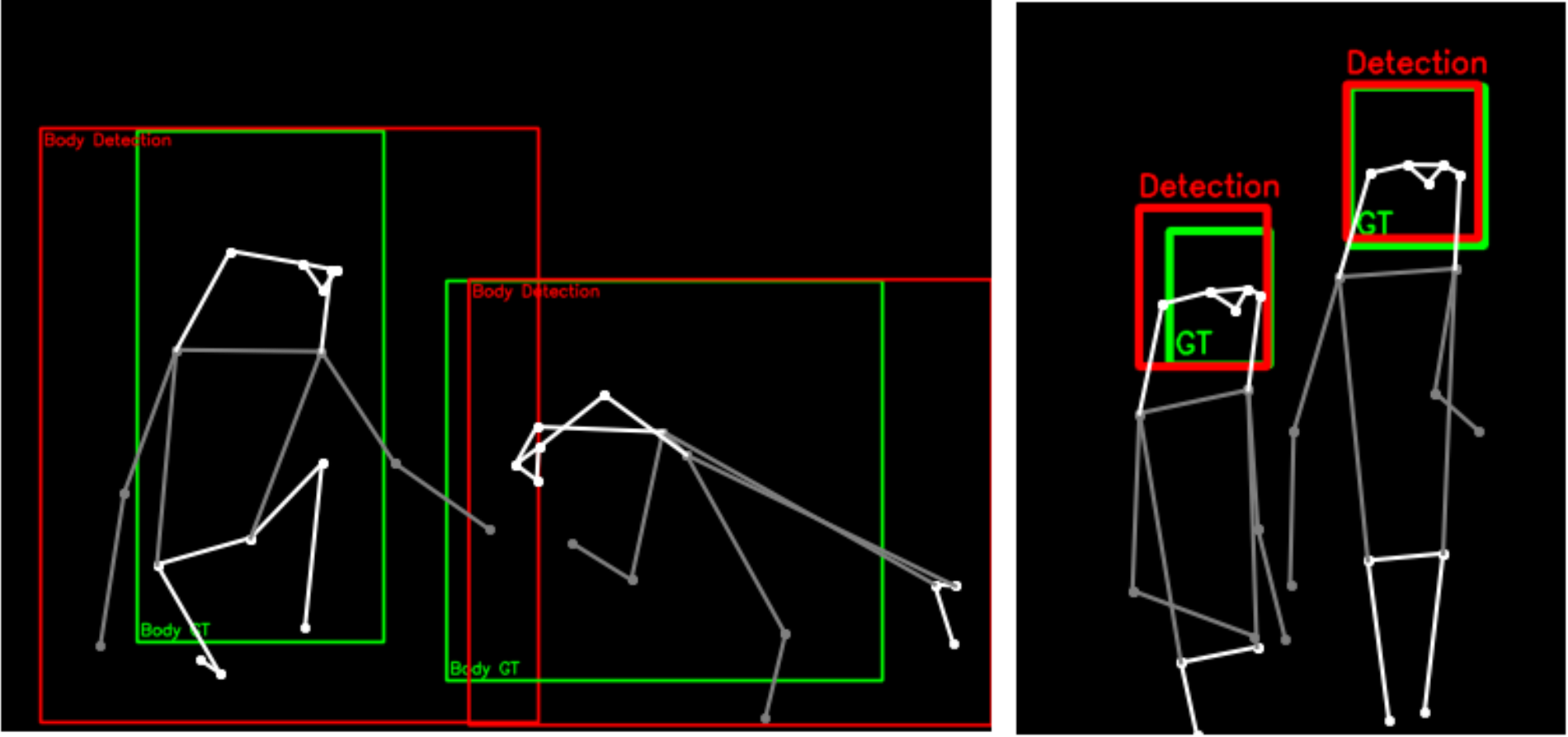}
    \caption{Comparison of inferences and ground-truth on RCPD. \textcolor{red}{Red}: detection, \textcolor{green}{green}: ground-truth. Gray traces refer to pose estimation results. Left: person, Right:face/head.}
    \label{fig:rcpd_iou_examples}
\end{figure}

For person detection, law enforcement experts tasked with labeling RCPD opted to minimize overlap between bounding boxes, thus partially excluding arms and legs when extremities were far from the torso and excluding parts of the torso for highly overlapping people. Regarding the face class, although labels are only provided for visible faces, the bounding box extends beyond the region from eyes to mouth, including the forehead and ears, but excluding the back of the head. Figure~\ref{fig:rcpd_iou_examples} showcases manually inspected detections from our methods, compared to the ground truth provided. Pose estimation is included as a reference since we cannot show the images.

Table~\ref{tab:results-detection-rcpd} summarizes our results for person detection on RCPD with the standard IoU, which may be misleading given the ground-truth characteristics. The metrics suggest our models are limited to coarse detections, with average precision around $30\%$ and $AP_{50}$ nearing $80\%$. However, our models are designed to capture the full extent of the person, which is different from the ground-truth. Furthermore, fine-tuning on BKPD consistently improved results for sexually explicit contexts (CSAI or adult), specially recalling more individuals. This optimization comes at a slight cost, as performance for non-explicit samples marginally decreased. This trade-off suggests that further refinement is needed to balance~performance.

\begin{table}[t]
    \centering
    \setlength\tabcolsep{0.5pt}
    \begin{tabular*}{\linewidth}{@{\extracolsep{\fill}} lccc|cccc|cccc}
    \toprule
    \multirow{2}{*}{\textbf{Method}} & 
    \multirow{2}{*}{\textbf{Domain}} & 
    \multirow{2}{*}{\textbf{Tune}} & &
    \multicolumn{4}{c|}{\textbf{\#People $=1$}} &
    \multicolumn{4}{c}{\textbf{\#People $\geq2$}} 
    \\ 
    
      & & & & AP   & AP$_{50}$ & AR & AR$_{50}$ 
            & AP   & AP$_{50}$ & AR & AR$_{50}$ \\ \midrule

    BKP-Assoc. & CSAI & -- &
               & \textbf{30.3}  & 66.1 & 44.3  & 76.7  
               & 27.3  & 60.6 & 38.7  & 69.1   \\ 
               
    YOLO-BKP   & CSAI & -- &
               & 29.6  & 65.1 & 46.0  & 79.5
               & \textbf{28.8}  & \textbf{63.3} & 41.0  & 73.8 
               \\ 
               
    BKP-Assoc. & CSAI & BKPD &
               & 30.2  & \textbf{67.6} & 44.1  & 77.8  
               & 27.7  & 62.9 & 38.4  & 70.4 \\

    YOLO-BKP   & CSAI & BKPD & 
               & 29.8  & 66.3 & \textbf{47.6}  & \textbf{81.4}  
               & 27.6  & \textbf{63.3} & \textbf{41.9}  & \textbf{77.3} 
               \\ \midrule 

    BKP-Assoc. & Porn & -- &
               & 34.3  & 78.7 & 50.5  & 89.4  
               & 23.3  & 58.5 & 37.3  & 72.3  
               \\ 
    YOLO-BKP   & Porn & -- &
               & 33.2  & 77.3 & 51.9  & 90.5 
               & \textbf{25.0}  & 59.9 & 39.5  & 75.7 
               \\ 
               
    BKP-Assoc. & Porn & BKPD &
               & \textbf{34.8}  & \textbf{81.9} & 50.7  & 91.0  
               & 22.2  & \textbf{62.3} & 36.9  & 76.3   
               \\ 
    YOLO-BKP   & Porn & BKPD & 
               & 34.1  & 80.1 & \textbf{52.5} & \textbf{92.5} 
               & 23.8  & 62.0 & \textbf{42.0}  & \textbf{81.4} 
               \\ \midrule 

    BKP-Assoc. & Other & -- &
               & 43.0  & 82.4 & \textbf{59.2}  & \textbf{92.4}  
               & \textbf{28.5}  & \textbf{68.0} & 40.6  & \textbf{77.2}
               \\
    YOLO-BKP   & Other & -- &
               & 41.2  & 80.4 & 57.8  & 90.7   
               & 28.1  & 65.7 & \textbf{41.8}  & 76.5   
               \\ 
               
    BKP-Assoc. & Other & BKPD &
               & \textbf{43.9}  & \textbf{84.2} & 58.6  & 91.8
               & 26.0  & 58.4 & 40.0  & 74.3 
               \\ 
    YOLO-BKP   & Other & BKPD & 
               & 41.1  & 79.9 & 56.7  & 90.2  
               & 24.7  & 60.3 & 40.7  & 73.5
               \\ 
    \bottomrule
    \end{tabular*}
    \caption{Detection summary of results with the standard IoU metric for the class person on RCPD.}
    \label{tab:results-detection-rcpd}
\end{table}

Examining Table~\ref{tab:results-detection-rcpd} horizontally reveals the impact of human interaction on performance. Adult pornography has the largest disparity between images with a single person and those with two or more people, up to 20 percentage points in $AP_{50}$. For images with a single person, the domain of CSAI has the lowest performance. As that could mean worse ground-truth discrepancies, we present on Table~\ref{tab:results-inner-detection-rcpd} precision and recall metrics calculated with the inner IoU, a measure employed in the association step of BPJDet~\cite{zhou2024bpjdet}. In our context, it is the ratio between (1) the intersection of prediction and ground truth and (2) the ground-truth area, i.e., it conveys whether the ground truth lies within prediction boundaries.

\begin{table}[t]
    \centering
    \setlength\tabcolsep{1.5pt}
    \begin{tabular}{lc|cccc|cccc }
    \toprule
      \multirow{2}{*}{\textbf{Method}} 
    & \multirow{2}{*}{\textbf{Domain}} 
    & \multicolumn{4}{c|}{\textbf{\#People $=1$}}
    & \multicolumn{4}{c}{\textbf{\#People $\geq 2$}}
    \\ 
        & &  AP$_{60}^{in}$   & AP$_{75}^{in}$  & AR$_{60}^{in}$ & AR$_{75}^{in}$  & 
              AP$_{60}^{in}$   & AP$_{75}^{in}$ & AR$_{60}^{in}$ & AR$_{75}^{in}$ \\ \midrule
    
    BKP-Assoc. & CSAI
               & \textbf{96.7}  & \textbf{94.4}   & 99.0 & 97.5 
               & \textbf{81.2}  & \textbf{80.5}   & \textbf{83.7} & \textbf{82.9}  \\ 
    YOLO-BKP   & CSAI
               & 94.1  & 91.6   & \textbf{99.2} & \textbf{98.1}
               & 81.0  & 80.0   & 83.4 & 82.5   \\ \midrule

    BKP-Assoc. & Porn
               & \textbf{99.2}  & \textbf{98.6}  & \textbf{100} & \textbf{99.5}  
               & 84.8  & 83.9  & 88.1 & 87.6     \\ 
    YOLO-BKP   & Porn
               & 97.9  & 95.3 & 99.5 & 97.5  
               & \textbf{86.2}  & \textbf{85.8} & \textbf{89.2} & \textbf{89.2}  \\ \midrule     
    BKP-Assoc. & Other
               & \textbf{98.6}  & \textbf{98.0}   & \textbf{100}  & \textbf{99.6} 
               & \textbf{89.6}  & \textbf{89.5}   & \textbf{94.1} & \textbf{94.1}   \\ 
    YOLO-BKP   & Other
               & 98.0  & 97.0  & 99.8 & 99.1 
               & 89.0  & 88.3  & 93.4 & 92.7   \\ 
    
    \bottomrule
    \end{tabular}
    \caption{Person detection results for models fine-tuned on BKPD, evaluated with inner IoU.}
    \label{tab:results-inner-detection-rcpd}
\end{table}

To understand Table~\ref{tab:results-inner-detection-rcpd} consider that $AP_{75}^{in}$ assigns a true positive when at least $75\%$ of the ground-truth is within prediction boundaries. The thresholds $\tau^{in} \in \{0.6, 0.75\}$ were selected based on BPJDet~\cite{zhou2024bpjdet}, which uses $\tau^{in}=0.6$ for association. We include $0.75$ as a more strict criterion for comparison. Even under a less strict evaluation, CSAI consistently yield worse results. This evaluation also shows the challenge of sexually explicit images with two or more people.

\begin{figure}[t]
    \centering
    \includegraphics[width=\linewidth]{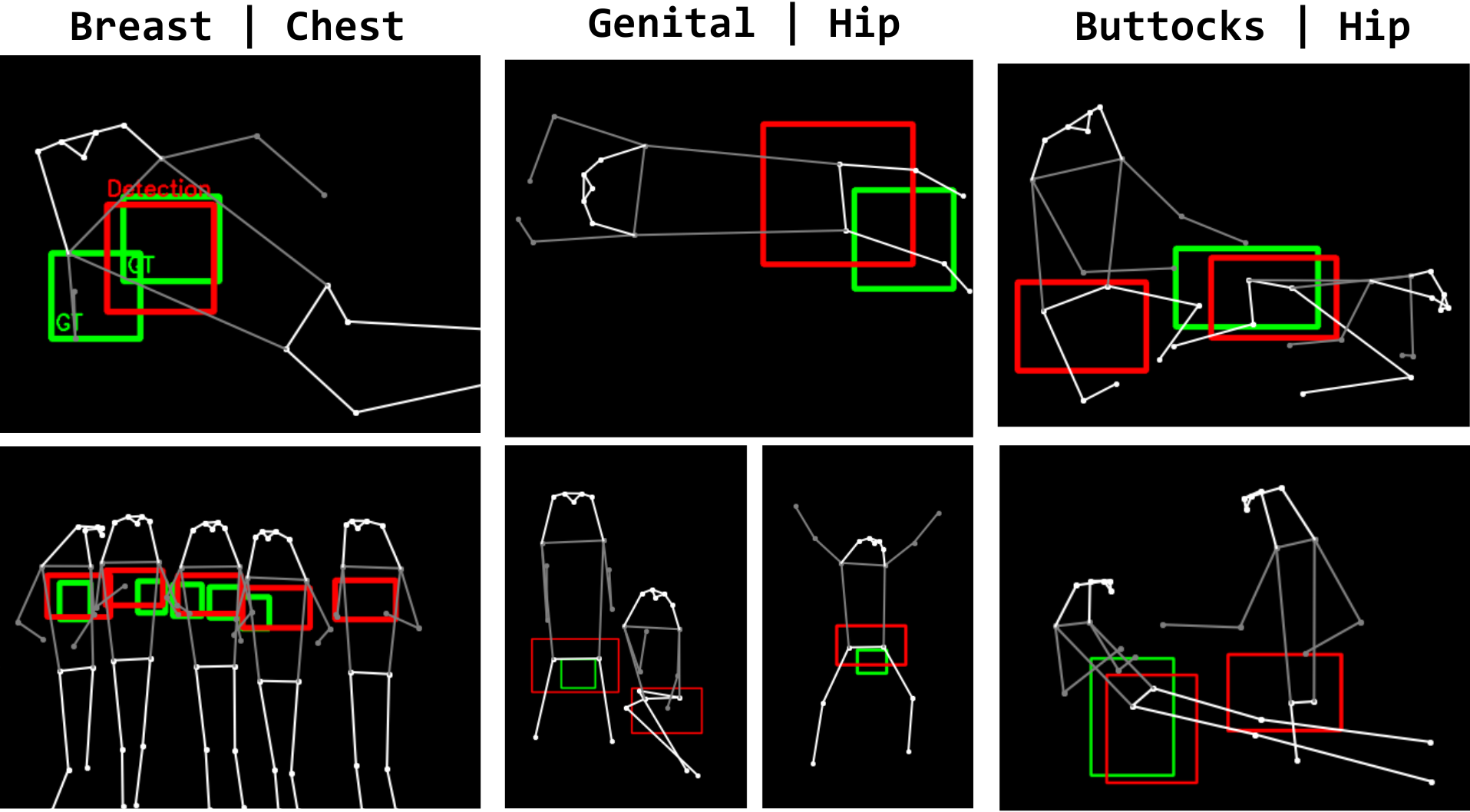}
    \caption{Example detections of private parts on RCPD. Detections in \textcolor{red}{red} and ground-truths in \textcolor{green}{green}.}
    \label{fig:rcpd_private_examples}
\end{figure}

Regarding the overlap between BKPD categories chest and hip with RCPD categories breast, genitalia, and buttocks, BKPD is designed to be domain-agnostic, locating regions of interest regardless of nudity or sexual explicitness, which does not correspond to RCPD labels. As illustrated in Figure~\ref{fig:rcpd_private_examples}, a substantial number of false positives are likely to emerge in a standard evaluation framework, even though the model is performing as expected. Furthermore, our models produce detections that adhere closely to the torso, partially excluding parts that extend beyond the body’s outline. 

To address the anticipated challenges, we employ the inner IoU metric, measuring whether RCPD labels are within the boundaries of our predictions. Furthermore, for each target, our evaluation will be restricted to samples that include the corresponding target. While it reduces the task's complexity, it enables meaningful result analysis. In Figure~\ref{fig:rcpd_private_examples}, at least one quantitative false positive occurs in each image, even under this less stringent experimental framework. 


\begin{table}[t]
    \centering
    \setlength\tabcolsep{0.5pt}
    \begin{tabular*}{\linewidth}{@{\extracolsep{\fill}} lcc|ccccc|cccc}
    \toprule
      \multirow{2}{*}{\textbf{Method}}  
    & \multirow{2}{*}{\textbf{Domain}} &   
    & \multicolumn{5}{c|}{\textbf{\#People $\leq1$}} 
    & \multicolumn{4}{c}{\textbf{\#People $\geq2$}}
    \\
    & & &  AR$_{50}^{in}$  
    &    AP$_{50}^{in}$   & JAP$_{50}^{in}$  & CA &
    &    AR$_{50}^{in}$ 
    &    AP$_{50}^{in}$   & JAP$_{50}^{in}$  & CA \\ \midrule
    & & &  \multicolumn{8}{c}{\textbf{Head \text{\textbar} Face}} \\ 
    \midrule
    BKP-Association & CSAI &
    & 99.0 & 97.9 & 97.9  & 100   &
    & 95.1 & 92.2 & 90.2  & 98.9  \\
    YOLO-BKP        & CSAI &
    & 98.4 & 95.5 & 95.2  & 98.8   &
    & 95.8 & 92.0 & 90.6  & 97.8  \\ \midrule
    BKP-Association & Porn &
    & 98.9 & 97.6 & 97.2  & 99.5   &
    & 96.2 & 92.2 & 91.4  & 99.3   \\
    YOLO-BKP        & Porn &
    & 99.5 & 98.2 & 97.7  & 98.4   &
    & 96.8 & 93.9 & 92.9  & 98.0  \\ \midrule

    & & &  \multicolumn{8}{c}{\textbf{Chest \text{\textbar} Breast}} \\ 
    \midrule
    BKP-Association & CSAI &
    & 77.7 & 66.1 & 65.9  & 99.6   &
    & 83.8 & 60.3 & 59.4  & 99.2  \\
    YOLO-BKP        & CSAI &
    & 84.4 & 73.7 & 73.2  & 99.7  &
    & 88.9 & 58.9 & 58.9  & 100  \\ \midrule
    BKP-Association & Porn &
    & 84.2 & 76.1 & 76.1  & 100   &
    & 89.1 & 68.5 & 68.5  & 100  \\
    YOLO-BKP        & Porn &
    & 87.0 & 77.7 & 77.7  & 100   &
    & 89.1 & 66.3 & 65.1  & 99.3  \\ \midrule

    & & &  \multicolumn{8}{c}{\textbf{Hip \text{\textbar} Genitalia}} \\ 
    \midrule
    BKP-Association & CSAI &
    & 83.3 & 78.9 & 78.8  & 99.5   &
    & 59.1 & 50.5 & 49.6  & 98.5  \\
    YOLO-BKP        & CSAI &
    & 88.0 & 82.4 & 82.3  & 99.5   &
    & 63.5 & 55.7 & 55.6  & 98.6  \\ \midrule
    BKP-Association & Porn &
    & 88.3 & 84.7 & 84.7  & 100   &
    & 73.9 & 59.8 & 57.2  & 96.1  \\
    YOLO-BKP        & Porn &
    & 91.0 & 87.0 & 86.9  & 99.2   &
    & 81.2 & 59.7 & 59.7  & 100  \\ \midrule

    & & &  \multicolumn{8}{c}{\textbf{Hip \text{\textbar} Buttocks}} \\ 
    \midrule
    BKP-Association & CSAI &
    & 82.0 & 78.7 & 75.0  & 96.9   &
    & 87.5 & 70.5 & 70.5  & 100  \\
    YOLO-BKP        & CSAI &
    & 87.2 & 80.6  & 80.6  & 100   &
    & 93.75 & 74.4 & 74.4  & 100   \\ \midrule
    BKP-Association & Porn &
    & 87.5 & 84.7 & 84.7  & 100   &
    & 100 & 83.4 & 83.4  & 100  \\
    YOLO-BKP        & Porn &
    & 100 & 80.0 & 80.0  & 100   &
    & 77.8 & 45.9 & 45.9  & 100  \\

    \bottomrule
    \end{tabular*}
    \caption[Detection and association performance of human parts from RCPD, evaluated with inner IoU.]{Detection and association performance of RCPD human parts, evaluated with inner IoU.}
    \label{tab:results-inner-association-rcpd}
\end{table}

Table~\ref{tab:results-inner-association-rcpd} summarizes results, revealing that YOLO-BKP demonstrated superior capability in learning novel targets and generalizing to the unseen domain of CSAI. This finding is particularly significant given the research focus on designing approaches robust to domain shifts. 
We highlight as the greatest issue in our study the presence of multiple humans in sexually explicit contexts, which affects detection performance across both adult pornography and CSAI.

A notable area of concern was the category genital, with the poorest performance. To understand false negatives in this category, we manually inspected the model's inferences --- without viewing the original images, in compliance with ethical constraints. Many false negatives occurred in cases involving extreme partial views of a person's body, such as close-ups on genitals or when most of the person's body was out-of-frame. The dataset labeled 73 out of the 82 false negatives as male, with the genital bounding boxes extending significantly farther from the torso. For the breast category, although few false positives were observed, the majority involved complex poses or intricate interactions, presenting notable challenges for detection. 

\section{Ethical Considerations}

This study involved an expert from the Federal Police of Brazil, who is solely responsible for handling any CSAI data. A significant challenge was balancing the need to scrutinize our results with the imperative to safeguard sensitive information. To enable qualitative assessment, we generated synthetic images containing only the skeletons of individuals as visual references for the depicted people (Figures \ref{fig:rcpd_private_examples} and \ref{fig:rcpd_iou_examples}), allowing us to evaluate specific detection targets without compromising safety. We emphasize that future research on CSAI should further pursue such efforts, as studies often avoid detailed analysis and rely solely on summary metrics.

Regarding our proposed dataset, we obtained samples from publicly available, peer-reviewed sources, following their established protocol. We intend to release our dataset upon request, subject to research-use agreement. Additionally, given its sensitive nature, with identifiable faces and sexually explicit content, BKPD depictions on this manuscript were blurred.
\section{Conclusion}
\label{sec:conclusion}

This work studies previously unexplored targets in the domain of CSAI, revealing important challenges in human-centric perception across the domains of safe images, adult pornography, and CSAI. The CSAI classification literature highlights a gap in addressing human-centric cues, such as posing and interactions, which law enforcement emphasizes but computer vision solutions overlook. To bridge this gap, we propose a novel dataset and two pipelines for simultaneous pose estimation and body parts detection, applicable to all human bodies regardless of nudity or demographics. 


We validated our method on widely adopted COCO-based benchmarks, achieving competitive performance. We further conducted cross-domain ablation studies using our proposed dataset, revealing key differences between safe images and adult pornography. In sexually explicit domains, regions such as the chest and hips proved more challenging to detect, as our method tends to localize detections near the torso. Finally, we conducted a case study in the CSAI domain using RCPD ground-truth annotations for person and private-part detection. Sensitive data were overseen by a Federal Police of Brazil expert, who performed random spot checks to ensure data integrity. Our findings suggest that even partial detections, while not fully capturing private parts, may still provide useful human-centered regions of interest.

Future directions for this study include research on higher-level human-centric tasks such as whole-body age estimation and interaction modeling for sexually explicit domains including scene graph construction and action recognition. 
Additionally, a systematic cross-domain evaluation should be conducted as our contribution was limited to revealing the challenges of human-centric perception tasks when transitioning from safe images to adult pornography and, ultimately, to CSAI.

\section*{Acknowledgements}
This work is partially funded by FAPESP through the Araceli Project (2023/12086-9), the CAPES Ph.D.~scholarship to C.~Laranjeira (88887.716349/2022-00), S.~Avila's funding by FAPESP (2023/12865-8, 2020/09838-0, 2013/08293-7), H.IAAC (01245.003479/2024-10) and CNPq (316489/2023-9), and J. A.~dos Santos' support by the Serrapilheira Institute (R-2011-37776).

\bibliographystyle{IEEEtran}
\bibliography{references}

\appendix
\subsection{BKPD Datasheet}
\label{FirstAppendix}

\definecolor{darkblue}{RGB}{46,25, 110}

\newcommand{\dssectionheader}[1]{%
   \noindent\framebox[\columnwidth]{%
      {\textbf{\textcolor{darkblue}{#1}}}
   }
}

\newcommand{\dsquestion}[1]{%
    {\noindent \textcolor{darkblue}{\textbf{#1}}}
}

\newcommand{\dsquestionex}[2]{%
    {\noindent \textcolor{darkblue}{\textbf{#1} #2}}
}

\newcommand{\dsanswer}[1]{%
   {\noindent #1 \medskip}
}


\dssectionheader{Motivation}

\dsquestionex{For what purpose was the dataset created?}{Was there a specific task in mind? Was there a specific gap that needed to be filled? Please provide a description.}

\dsanswer{
The main purpose is to approximate the domain of Child Sexual Abuse Imagery (CSAI) leveraging legal images related to sub-tasks from the target domain. Specifically, we include safe images of people from varying age groups, along with sexually explicit images of adults, sourced from publicly available and peer-reviewed datasets.

Regarding the annotated targets, the intended tasks are pose estimation and detection of persons and selected body parts associated with CSAI, namely: head, chest, hip and hands. Our dataset allows to individually target these tasks, as well as predicting associated targets for each individual.
}

\dsquestion{Who created this dataset (e.g., which team, research group) and on behalf of which entity (e.g., company, institution, organization)?}

\dsanswer{
The dataset was created within Project Araceli: Inteligência Artificial no Combate ao Abuso Sexual Infantil (\textit{Artificial Intelligence in the Fight Against Child Sexual Abuse}). The project is a joint effort from several universities in Brazil, in partnership with Brazil's Federal Police. Specifically, the dataset collection was part of the PhD research of Camila Laranjeira da Silva at the Federal University of Minas Gerais. 
}

\dsquestionex{Who funded the creation of the dataset?}{If there is an associated grant, please provide the name of the grantor and the grant name and number.}

\dsanswer{
Project Araceli is funded by the Brazilian agency FAPESP  under grant 23/12086-9. Camila's PhD was funded by a
CAPES Ph.D.~scholarship (88887.716349/2022-00). This work was also partially supported by Serrapilheira Institute under grant Serra–R-2011-37776.
}

\dsquestion{Any other comments?}

\bigskip
\dssectionheader{Composition}

\dsquestionex{What do the instances that comprise the dataset represent (e.g., documents, photos, people, countries)?}{ Are there multiple types of instances (e.g., movies, users, and ratings; people and interactions between them; nodes and edges)? Please provide a description.}

\dsanswer{
The instances comprise images depicting one or more individuals, along with an annotation file with skeleton coordinates and bounding boxes for each image and each depicted person.
}

\dsquestion{How many instances are there in total (of each type, if appropriate)?}

\dsanswer{
A total of $1,985$ images and the corresponding annotation files. $1,002$ images are sexually explicit, and $983$ are safe images of people. 
}

\dsquestionex{Does the dataset contain all possible instances or is it a sample (not necessarily random) of instances from a larger set?}{ If the dataset is a sample, then what is the larger set? Is the sample representative of the larger set (e.g., geographic coverage)? If so, please describe how this representativeness was validated/verified. If it is not representative of the larger set, please describe why not (e.g., to cover a more diverse range of instances, because instances were withheld or unavailable).}

\dsanswer{
BKPD is a small dataset representing two CSAI-related sub-tasks, namely safe images of people from varying ages and sexually explicit images of adults. The dataset is not entirely representative of the larger set, as it does not comprise all variations both from safe and sexually explicit contexts involving people.
The samples are sourced from other datasets, respectively Open Images\footnote{\url{https://storage.googleapis.com/openimages/web/index.html}} and the Sexual Organ Dataset (SOD)\footnote{Tabone, André, et al. "Pornographic content classification using deep-learning." Proceedings of the 21st ACM symposium on document engineering. 2021.}. We acknowledge that Open Images is limited in identifying age diversity, as it only provides binary young/adult labels. Additionally, SOD presents several bias, being a collection of pornographic images from mainstream websites focusing on the female body.
}

\dsquestionex{What data does each instance consist of? “Raw” data (e.g., unprocessed text or images) or features?}{In either case, please provide a description.}

\dsanswer{
Each instance is a raw image.
}

\dsquestionex{Is there a label or target associated with each instance?}{If so, please provide a description.}

\dsanswer{
The annotation file for each image contains coordinates for each skeleton target and the bounding boxes for each mapped category. Targets are associated per individual.
}

\dsquestionex{Is any information missing from individual instances?}{If so, please provide a description, explaining why this information is missing (e.g., because it was unavailable). This does not include intentionally removed information, but might include, e.g., redacted text.}

\dsanswer{
No.
}

\dsquestionex{Are relationships between individual instances made explicit (e.g., users’ movie ratings, social network links)?}{If so, please describe how these relationships are made explicit.}

\dsanswer{
There is no explicit relationship between samples, except the dataset from which they were sourced. Samples are marked as safe or sexually explicit.
}

\dsquestionex{Are there recommended data splits (e.g., training, development/validation, testing)?}{If so, please provide a description of these splits, explaining the rationale behind them.}

\dsanswer{
BKPD is intended primarily for testing purposes, thus we do not recommend data splits.
}

\dsquestionex{Are there any errors, sources of noise, or redundancies in the dataset?}{If so, please provide a description.}

\dsanswer{
The annotations of skeleton coordinates and body part bounding boxes are manually selected or drawn by human annotators. This process is inherently subject to a certain level of noise.
}

\dsquestionex{Is the dataset self-contained, or does it link to or otherwise rely on external resources (e.g., websites, tweets, other datasets)?}{If it links to or relies on external resources, a) are there guarantees that they will exist, and remain constant, over time; b) are there official archival versions of the complete dataset (i.e., including the external resources as they existed at the time the dataset was created); c) are there any restrictions (e.g., licenses, fees) associated with any of the external resources that might apply to a future user? Please provide descriptions of all external resources and any restrictions associated with them, as well as links or other access points, as appropriate.}

\dsanswer{
BKPD is self-contained.
}

\dsquestionex{Does the dataset contain data that might be considered confidential (e.g., data that is protected by legal privilege or by doctor-patient confidentiality, data that includes the content of individuals non-public communications)?}{If so, please provide a description.}

\dsanswer{
The images were sourced from publicly available, peer-reviewed, datasets. The original collection process, described in each paper, refers to publicly available images.
}

\dsquestionex{Does the dataset contain data that, if viewed directly, might be offensive, insulting, threatening, or might otherwise cause anxiety?}{If so, please describe why.}

\dsanswer{
Yes. Adult pornography is considered sensitive media, and thus, viewer discretion is advised, as it comprises sexually explicit depictions.
}

\dsquestionex{Does the dataset relate to people?}{If not, you may skip the remaining questions in this section.}

\dsanswer{
Yes.
}

\dsquestionex{Does the dataset identify any subpopulations (e.g., by age, gender)?}{If so, please describe how these subpopulations are identified and provide a description of their respective distributions within the dataset.}

\dsanswer{
The provided annotations do not allow the identification of subpopulations. However, visualizing raw images allow inferring perceived attributes. Additionally, since filenames are kept as per the original datasets, one can refer to age and gender labels from Open Images, and gender-assigned bounding boxes from SOD.
}

\dsquestionex{Is it possible to identify individuals (i.e., one or more natural persons), either directly or indirectly (i.e., in combination with other data) from the dataset?}{If so, please describe how.}

\dsanswer{
Since the dataset contains raw images of people, often with visible faces, techniques such as reverse image search may allow the identification of individuals.
}

\dsquestionex{Does the dataset contain data that might be considered sensitive in any way (e.g., data that reveals racial or ethnic origins, sexual orientations, religious beliefs, political opinions or union memberships, or locations; financial or health data; biometric or genetic data; forms of government identification, such as social security numbers; criminal history)?}{If so, please provide a description.}

\dsanswer{
Yes. Adult pornography is a sensitive media, revealing aspects related to sexuality. More generally, images of people with visible faces characterize biometric data.
}

\dsquestion{Any other comments?}

\bigskip
\dssectionheader{Collection Process}

\dsquestionex{How was the data associated with each instance acquired?}{Was the data directly observable (e.g., raw text, movie ratings), reported by subjects (e.g., survey responses), or indirectly inferred/derived from other data (e.g., part-of-speech tags, model-based guesses for age or language)? If data was reported by subjects or indirectly inferred/derived from other data, was the data validated/verified? If so, please describe how.}

\dsanswer{
BKPD is a collection from publicly available, peer-reviewed, datasets. Our data collection process was random selection for sexually explicit images sourced from SOD, and random selection of images with age and gender labels (girl, boy, man, woman) for Open Images. For details on the original data collection of images, please refer to the documentation of each dataset.
}

\dsquestionex{What mechanisms or procedures were used to collect the data (e.g., hardware apparatus or sensor, manual human curation, software program, software API)?}{How were these mechanisms or procedures validated?}

\dsanswer{
Images were downloaded from each dataset's original source.
}

\dsquestion{If the dataset is a sample from a larger set, what was the sampling strategy (e.g., deterministic, probabilistic with specific sampling probabilities)?}

\dsanswer{
Images were randomly sampled from each source dataset.  
}

\dsquestion{Who was involved in the data collection process (e.g., students, crowdworkers, contractors) and how were they compensated (e.g., how much were crowdworkers paid)?}

\dsanswer{
No financial compensation was involved, beyond research grants and scholarships. Data collection and annotation was conducted by the authors of the present study.
}

\dsquestionex{Over what timeframe was the data collected? Does this timeframe match the creation timeframe of the data associated with the instances (e.g., recent crawl of old news articles)?}{If not, please describe the timeframe in which the data associated with the instances was created.}

\dsanswer{
The data was collected in january 2024, downloading images from the source datasets. For details on the collection timeframe of the original images, please refer to the documentation of each dataset.
}

\dsquestionex{Were any ethical review processes conducted (e.g., by an institutional review board)?}{If so, please provide a description of these review processes, including the outcomes, as well as a link or other access point to any supporting documentation.}

\dsanswer{
No. The images come from publicly available, peer-reviewed, datasets. 
}

\dsquestionex{Does the dataset relate to people?}{If not, you may skip the remaining questions in this section.}

\dsanswer{
Yes.
}

\dsquestion{Did you collect the data from the individuals in question directly, or obtain it via third parties or other sources (e.g., websites)?}

\dsanswer{
Images were obtained from publicly available sources.
}

\dsquestionex{Were the individuals in question notified about the data collection?}{If so, please describe (or show with screenshots or other information) how notice was provided, and provide a link or other access point to, or otherwise reproduce, the exact language of the notification itself.}

\dsanswer{
No. Individuals were not notified about the data collection.
}

\dsquestionex{Did the individuals in question consent to the collection and use of their data?}{If so, please describe (or show with screenshots or other information) how consent was requested and provided, and provide a link or other access point to, or otherwise reproduce, the exact language to which the individuals consented.}

\dsanswer{
The authors responsible for Open Images report that collected images are under a CC BY 2.0 license. Conversely, the authors responsible for SOD do not report license status.
}

\dsquestionex{If consent was obtained, were the consenting individuals provided with a mechanism to revoke their consent in the future or for certain uses?}{If so, please provide a description, as well as a link or other access point to the mechanism (if appropriate).}

\dsanswer{
Direct consent was not obtained.
}

\dsquestionex{Has an analysis of the potential impact of the dataset and its use on data subjects (e.g., a data protection impact analysis) been conducted?}{If so, please provide a description of this analysis, including the outcomes, as well as a link or other access point to any supporting documentation.}

\dsanswer{
No such analysis was conducted.
}

\dsquestion{Any other comments?}

\dsanswer{
}

\bigskip
\dssectionheader{Preprocessing/cleaning/labeling}

\dsquestionex{Was any preprocessing/cleaning/labeling of the data done (e.g., discretization or bucketing, tokenization, part-of-speech tagging, SIFT feature extraction, removal of instances, processing of missing values)?}{If so, please provide a description. If not, you may skip the remainder of the questions in this section.}

\dsanswer{
The images were labelled with skeleton coordinates and bounding boxes for persons and selected body parts. 
}

\dsquestionex{Was the “raw” data saved in addition to the preprocessed/cleaned/labeled data (e.g., to support unanticipated future uses)?}{If so, please provide a link or other access point to the “raw” data.}

\dsanswer{
The labels are provided separately from the raw data, i.e., raw images.
}

\dsquestionex{Is the software used to preprocess/clean/label the instances available?}{If so, please provide a link or other access point.}

\dsanswer{
The labels were produced with the open-sourced tool Label Studio~\footnote{\url{https://labelstud.io}}.
}

\dsquestion{Any other comments?}

\dsanswer{
}

\bigskip
\dssectionheader{Uses}

\dsquestionex{Has the dataset been used for any tasks already?}{If so, please provide a description.}

\dsanswer{
Yes. The dataset was published along with experiments for pose estimation, body part detection, and target association per individual.
}

\dsquestionex{Is there a repository that links to any or all papers or systems that use the dataset?}{If so, please provide a link or other access point.}

\dsanswer{
No. The dataset was only used a single time, by the authors themselves.
}

\dsquestion{What (other) tasks could the dataset be used for?}

\dsanswer{
The dataset could be used for adult pornography classification, as the filenames include the categories safe or sexually explicit. 
}

\dsquestionex{Is there anything about the composition of the dataset or the way it was collected and preprocessed/cleaned/labeled that might impact future uses?}{For example, is there anything that a future user might need to know to avoid uses that could result in unfair treatment of individuals or groups (e.g., stereotyping, quality of service issues) or other undesirable harms (e.g., financial harms, legal risks) If so, please provide a description. Is there anything a future user could do to mitigate these undesirable harms?}

\dsanswer{
Adult pornography datasets often comprise gender-assigned labels, distinguishing human bodies perceived as male or female. BKPD was purposefully labeled with gender neutral labels (e.g., chest, hip), as they are common to all human bodies.
}

\dsquestionex{Are there tasks for which the dataset should not be used?}{If so, please provide a description.}

\dsanswer{
Dealing with sensitive media inherently invites a careful approach. As a general recommendation, when using our datasets, the benefits should outweigh the costs.
}

\dsquestion{Any other comments?}

\bigskip
\dssectionheader{Distribution}

\dsquestionex{Will the dataset be distributed to third parties outside of the entity (e.g., company, institution, organization) on behalf of which the dataset was created?}{If so, please provide a description.}

\dsanswer{
Yes. The dataset will be available upon request, subject to research-use agreement.
}

\dsquestionex{How will the dataset will be distributed (e.g., tarball on website, API, GitHub)}{Does the dataset have a digital object identifier (DOI)?}

\dsanswer{
BKPD will be distributed through Zenodo~\footnote{\url{https://zenodo.org}}, which provides archiving and DOI generation.
}

\dsquestion{When will the dataset be distributed?}

\dsanswer{
Upon paper acceptance.
}

\dsquestionex{Will the dataset be distributed under a copyright or other intellectual property (IP) license, and/or under applicable terms of use (ToU)?}{If so, please describe this license and/or ToU, and provide a link or other access point to, or otherwise reproduce, any relevant licensing terms or ToU, as well as any fees associated with these restrictions.}

\dsanswer{
Dataset access will require users to register their intended use and institutional affiliation, and to agree that the data will be used solely for non-commercial research purposes.
}

\dsquestionex{Have any third parties imposed IP-based or other restrictions on the data associated with the instances?}{If so, please describe these restrictions, and provide a link or other access point to, or otherwise reproduce, any relevant licensing terms, as well as any fees associated with these restrictions.}

\dsanswer{
No.
}

\dsquestionex{Do any export controls or other regulatory restrictions apply to the dataset or to individual instances?}{If so, please describe these restrictions, and provide a link or other access point to, or otherwise reproduce, any supporting documentation.}

\dsanswer{
No.
}

\dsquestion{Any other comments?}

\dsanswer{
}

\bigskip
\dssectionheader{Maintenance}

\dsquestion{Who will be supporting/hosting/maintaining the dataset?}

\dsanswer{
The first author, Camila Laranjeira da Silva.
}

\dsquestion{How can the owner/curator/manager of the dataset be contacted (e.g., email address)?}

\dsanswer{
The owner can be contacted through the e-mail mila.laranjeira@gmail.com.
}

\dsquestionex{Is there an erratum?}{If so, please provide a link or other access point.}

\dsanswer{
No,
}

\dsquestionex{Will the dataset be updated (e.g., to correct labeling errors, add new instances, delete instances)?}{If so, please describe how often, by whom, and how updates will be communicated to users (e.g., mailing list, GitHub)?}

\dsanswer{
No.
}

\dsquestionex{If the dataset relates to people, are there applicable limits on the retention of the data associated with the instances (e.g., were individuals in question told that their data would be retained for a fixed period of time and then deleted)?}{If so, please describe these limits and explain how they will be enforced.}

\dsanswer{
No.
}

\dsquestionex{Will older versions of the dataset continue to be supported/hosted/maintained?}{If so, please describe how. If not, please describe how its obsolescence will be communicated to users.}

\dsanswer{
The dataset will not be updated with subsequent versions.
}

\dsquestionex{If others want to extend/augment/build on/contribute to the dataset, is there a mechanism for them to do so?}{If so, please provide a description. Will these contributions be validated/verified? If so, please describe how. If not, why not? Is there a process for communicating/distributing these contributions to other users? If so, please provide a description.}

\dsanswer{
No. Any contributions may be independently distributed.
}

\dsquestion{Any other comments?}

\dsanswer{
}


\end{document}